%% file: main.tex
\documentclass[10pt,twocolumn,letterpaper]{article}

\usepackage{booktabs}

\usepackage{iccv}
\usepackage{times}
\usepackage{epsfig}
\usepackage{graphicx}
\usepackage{amsmath}
\usepackage{amssymb}
\usepackage{color}
\usepackage{caption}
\usepackage{mathtools}
\usepackage{multirow}
\usepackage{times}
\usepackage{amsmath}
\usepackage{enumitem}
\usepackage{subcaption}

\newcommand{\f}[1]{f_\text{#1}}
\newcommand{\loss}[1]{L_\text{#1}}
\newcommand{\norm}[1]{\lVert{#1}\rVert}
\newcommand{\tabspace}{\hspace{6mm}}

\usepackage[pagebackref=true,breaklinks=true,letterpaper=true,colorlinks,bookmarks=false]{hyperref}

\iccvfinalcopy

\ificcvfinal\fi
\begin{document}

\title{Predicting 3D Human Dynamics from Video}

\author{Jason Y. Zhang, Panna Felsen, Angjoo Kanazawa, Jitendra Malik\\
University of California, Berkeley\\
\tt{\small \{zhang.j, panna, kanazawa, malik\}@eecs.berkeley.edu}
}

\twocolumn[{%
\renewcommand\twocolumn[1][]{#1}%
\maketitle

\begin{center}
    \newcommand{\teaserwidth}{\textwidth}
    \vspace{-3mm}
    \centerline{
    \includegraphics[width=\teaserwidth]{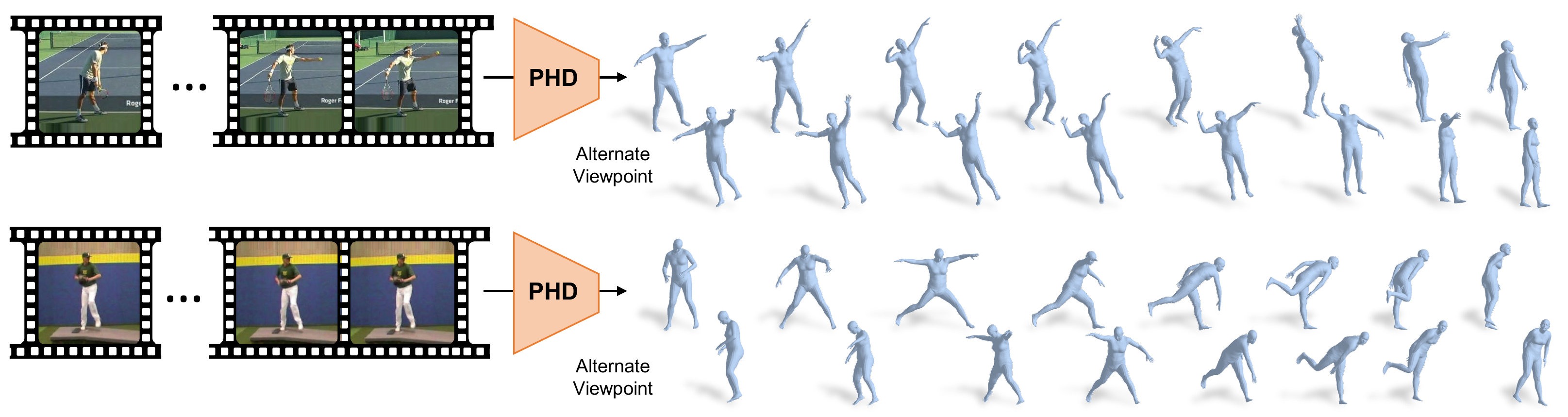}
     }

    \captionof{figure}{\textbf{Autoregressive prediction of human 3D motion from video.} We present Predicting Human Dynamics (PHD), a neural auto-regressive framework that takes past video frames as input to predict the motion of a 3D human body model. As shown, PHD takes in a video sequence of a person and predicts the future 3D human motion. We show the predictions from two different viewpoints.}
    \vspace{-2mm}
	\label{fig:teaser}

\end{center}%
}]

\begin{abstract}
\vspace{-4.2mm}
\input{phd_0_abstract.tex}
\end{abstract}

\input{phd_1_introduction.tex}
\input{phd_2_related.tex}
\input{phd_3_method.tex}

\input{phd_4_eval.tex}

\input{phd_5_conclusion.tex}

{\small
\bibliographystyle{ieee_fullname}
\bibliography{egbib}
}

\section{Appendix}
\input{phd_6_appendix.tex}

\end{document}

%% file: phd_0_abstract.tex
Given a video of a person in action, we can easily guess the 3D future motion of the person. In this work, we present perhaps the first approach for predicting a future 3D mesh model sequence of a person from past video input. We do this for periodic motions such as walking and also actions like bowling and squatting seen in sports or workout videos. While there has been a surge of future prediction problems in computer vision, most approaches predict 3D future from 3D past or 2D future from 2D past inputs. In this work, we focus on the problem of predicting 3D future motion from past image sequences, which has a plethora of practical applications in autonomous systems that must operate safely around people from visual inputs. Inspired by the success of autoregressive models in language modeling tasks, we learn an intermediate latent space on which we predict the future. This effectively facilitates autoregressive predictions when the input differs from the output domain. Our approach can be trained on video sequences obtained in-the-wild without 3D ground truth labels. The project website with videos can be found at \href{https://jasonyzhang.com/phd}{https://jasonyzhang.com/phd}.

%% file: phd_1_introduction.tex
\section{Introduction}
\label{sec:intro}

\begin{figure*}[t]
  \centering
  \includegraphics[width=\textwidth]{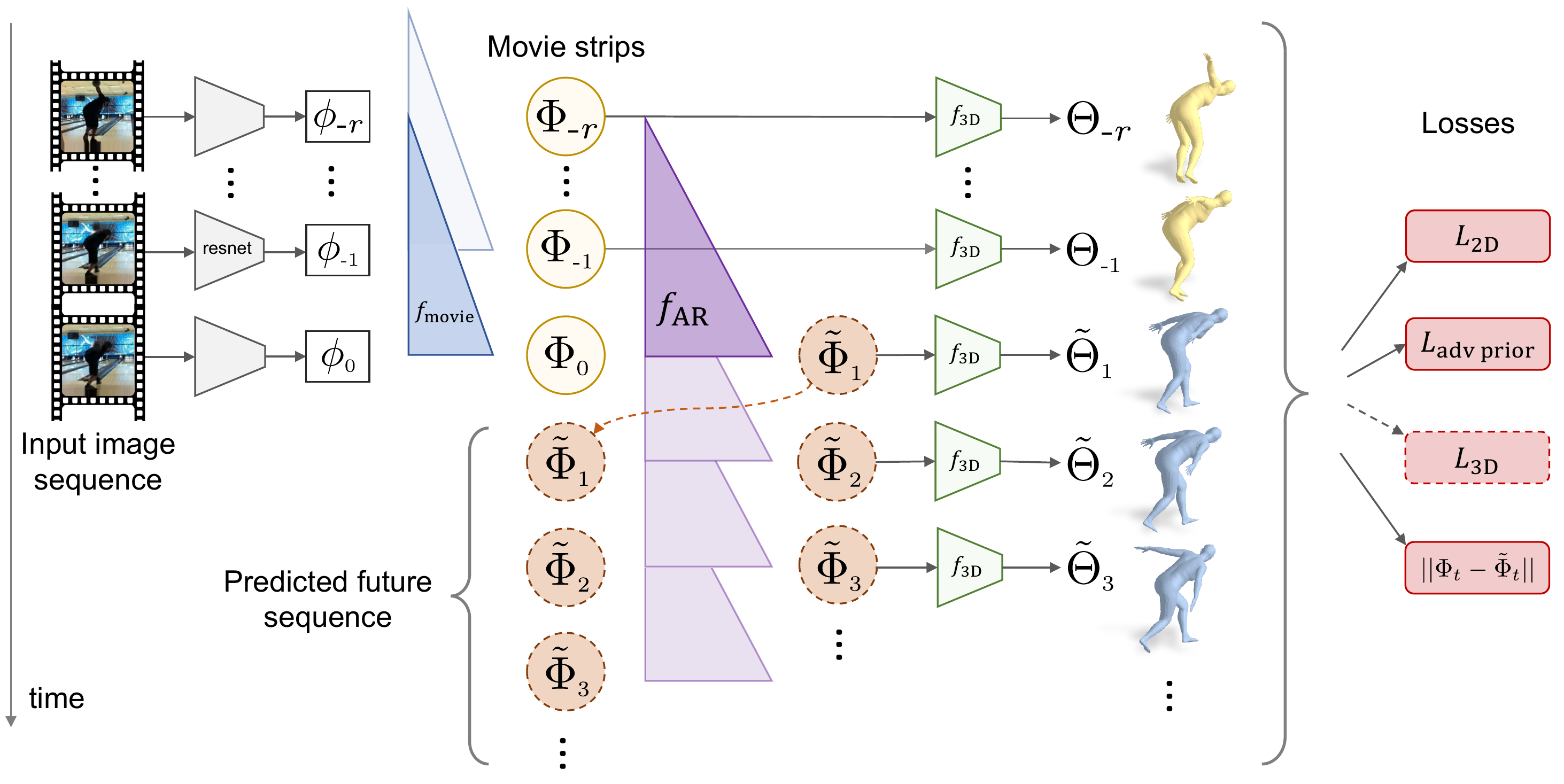}
  \vspace{-6mm}
  \caption{{{\bf Overview of Predicting Human Dynamics (PHD).} From a video sequence of a person, we predict the future 3D human mesh. We first extract per-image features $\phi_t$. We then train a causal temporal encoder $\f{movie}$ that learns a ``movie strip," a latent representation of 3D human dynamics $\Phi_t$ that captures the historical context leading up to time $t$. We also train a 3D regressor $\f{3D}$ that can read out the 3D mesh from the movie strip. In this latent space, we train an autoregressive $\f{AR}$ which takes the past movie strips to predict the future movie strips, thereby capturing the human dynamics.
  We include a 3D loss when 3D annotations are available. Otherwise, we train in a weakly-supervised manner on in-the-wild videos that lack 3D annotations using 2D re-projection error and an adversarial prior.
  	At test time, conditioned on half a second of video, PHD produces plausible human motion seconds into the future.
  }}
  \vspace{-3mm}
  \label{fig:overview}
\end{figure*}

Consider the video sequences in Figure \ref{fig:teaser}. Given the past frames of action, we can easily imagine the future motion of the athletes, whether hitting a tennis serve or pitching a baseball. In this paper, we consider this problem of predicting the motion of the person as a 3D mesh given some past image sequences. We propose a learning-based framework that given past frames can successively predict the future 3D mesh of the person in an autoregressive manner. Our model is trained on videos obtained in-the-wild without ground truth 3D annotations. 

Learning generative models of sequences has a long tradition, particularly in language \cite{bengio2003neural,graves2013generating} and speech generation \cite{oord2016wavenet}.
Our approach is the first counterpart to these approaches for 3D mesh motion generation from video. While there has been much interest in future prediction from video, most approaches focus on predicting 2D components from video such as 2D keypoints \cite{pos_iccv2017,yan2018mt}, flow \cite{walker2016uncertain}, or pixels \cite{denton2018stochastic, finn2016unsupervised}. 
On the other hand, the several works that predict 3D human motion all take past 3D skeleton sequences as input obtained from motion capture data.
To our knowledge, no previous approach explores the problem of 3D human motion prediction from video. 3D is a natural space of motion prediction with many practical applications such as human-robot interaction, where autonomous systems such as self-driving cars or drones must operate safely around people from visual inputs in-the-wild. Our approach is trainable on videos without 3D annotations, providing a more abundant and natural source of information than 3D motion sequences obtained from a motion capture studio.

We introduce an autoregressive architecture for this task following recent successes in convolutional autoregressive sequence modeling approaches \cite{bai2018empirical,oord2016wavenet}. A challenge introduced by the problem of 3D motion prediction from video is that the space of input (image frames) and that of output (3D meshes) are different. Following the advances in the language-modeling literature, we remedy this issue by learning a shared latent representation in which we can apply the future prediction model in an autoregressive manner.

We build on our previous work that learns a latent representation for 3D human dynamics $\Phi_i$ from the temporal context of image frames \cite{kanazawa19hmmr}. We modify the approach so that the convolutional model has a causal structure, where only past temporal context is utilized. Once we learn a causal latent representation for 3D human motion from video, we train an autoregressive prediction model in this latent space. While our previous work has demonstrated one-step future  prediction, it can only take a single image as an input and requires a separate future hallucinator for every time step. In contrast, in this paper, we learn an autoregressive model which can recurrently predict a longer range future (arbitrarily many frames versus 1 frame) that is more stable by taking advantage of a sequence of past image frames. To our knowledge, our work is the first study on predicting 3D human motion from image sequences. We demonstrate our approach on the Human3.6M dataset \cite{Human36m:2014} and the in-the-wild Penn Action dataset \cite{Zhang2013Penn}.

%% file: phd_2_related.tex
\section{Related Work}
\label{sec:related}

\noindent\textbf{Generative Modeling of Sequences.}
There has long been an interest in generative models of sequences in language and speech.
Modern deep learning-based approaches began with recurrent neural networks based models \cite{graves2013generating,sutskever2014sequence}. 
Feed-forward models with convolutional layers are also used for sequence modeling tasks, such as image generation \cite{van2016conditional} and audio waveform generation \cite{oord2016wavenet}. Recent studies suggest that these feed-forward models can outperform recurrent networks \cite{bai2018empirical} or do equivalently  \cite{miller2018recurrent}, while being parallelizable and easier to train with stable gradients. In this work, we also use feed-forward convolutional layers for our autoregressive future prediction model. 

\noindent\textbf{Visual Prediction.}
There are a number of methods that predict the future from video or images. 
Ryoo \cite{ryoo2011human} predicts future human activity classes from a video input. 
Kitani \etal \cite{kitani2012activity} predict possible trajectories of a person in the image from surveillance footage. \cite{kooij2014context} predict paths of pedestrians from a stereo camera on a car. \cite{koppula2016anticipating} anticipate action trajectories in a human-robot interaction from RGB-D videos. More recent deep learning-based approaches explore predicting denser 2D outputs such as raw pixels \cite{van2016conditional,finn2016unsupervised,denton2017unsupervised,denton2018stochastic}, 2D flow fields \cite{walker2016uncertain}, or more structured outputs like 2D human pose \cite{pos_iccv2017,yan2018mt}.

There are also approaches that from a single image predict future in the form of object dynamics \cite{fouhey2014predicting}, object trajectories \cite{walker2014patch}, flow \cite{pintea2014deja,walker2015dense,Prediction-ECCV-2018,gao2018im2flow}, a difference image \cite{xue2016visual}, action categories \cite{vu2014predicting}, or image representations \cite{vondrick2016anticipating}.
All approaches predict future in 2D domains or categories from input video. In this work, we propose a framework that predicts 3D motions from video inputs.

\noindent\textbf{3D Pose from Video.}
There has been much progress in recovering 3D human pose from RGB video. Common pose representations include 3D skeletons \cite{ martinez2017_3dbaseline, dabral2017_tpnet, VNect_SIGGRAPH2017, Mehta2018XNectRM, pavllo2019videopose3d} and meshes \cite{arnab2019exploiting, doersch2019sim2real_3dpose, kanazawa19hmmr, tung2017self}. We build on our previous weakly-supervised, mesh-based method that can take advantage of large-scale Internet videos without ground truth annotations \cite{kanazawa19hmmr}. While mesh-based methods do not necessarily have the lowest error on Human3.6M, recent work suggests that performance on Human3.6M does not correlate very well with performance on challenging in-the-wild video datasets such as 3DPW \cite{vonMarcard2018, kanazawa19hmmr}.

\noindent\textbf{3D to 3D Human Motion Prediction.}
Modeling the dynamics of humans has long-standing interest \cite{bregler1997learning}. Earlier works model the synthesis of human motion using techniques such as Hidden Markov Models \cite{brand2000style}, linear dynamical systems \cite{pavlovic2001learning}, bilinear spatiotemporal basis models \cite{akhter2012bilinear}, and Gaussian process latent variable models \cite{urtasun2008topologically, wang2008gaussian} and other variants \cite{hsu2005style,wang2007multifactor}. More recently, there are deep learning-based approaches that use recurrent neural networks (RNNs) to predict 3D future human motion from past 3D human skeletons \cite{fragkiadaki2015recurrent,jain2016structural,butepage2017deep,li2018auto,villegas2018neural}. All of these approaches operate in the domain where the inputs are 3D past motion capture sequences.  In contrast, our work predicts future 3D human motion from past 2D video inputs. 

\noindent\textbf{3D Future Prediction from Single Image.}
More related to our setting is that of Chao \etal
\cite{chao2017forecasting}, who from a single image, predict 2D future pose sequences from which the 3D pose can be estimated. While this approach does produce 3D skeletal human motion sequences from a single image, the prediction happens in the 2D to 2D domain. More recently, our previous \cite{kanazawa19hmmr} presents an approach that can predict a single future 3D human mesh from a single image by predicting the latent representation which can be used to get the future 3D human mesh. While this approach requires learning a separate future predictor for every time step, we propose a single autoregressive model that can be reused to successively predict the future 3D meshes.

%% file: phd_3_method.tex
\section{Approach}
\label{sec:method}

Our goal is to predict the future 3D mesh sequence of a human given past image sequences. Specifically, our input is a set of past image frames of a video $V = \{I_{t}, I_{t - 1}, ... I_{t-N}\}$, and our output is a future sequence of 3D human meshes $\Theta = \{\Theta_{t+1}, \Theta_{t+2},  \dots, \Theta_{t+T}\}$. We represent the future 3D mesh as $\Theta = [\theta, \beta]$ consisting of pose parameters $\theta$ and shape parameters $\beta$.

We propose Predicting Human Dynamics (PHD), a neural autoregressive network for predicting human 3D mesh sequences from video. Our network is divided into two components: one that learns a latent representation of 3D human motion from video, and another that learns an autoregressive model of the latent representation from which the 3D human prediction may be recovered. Figure \ref{fig:overview} shows an overview of the model.
For the first part, we build upon our recent work \cite{kanazawa19hmmr} which learns a latent representation of 3D human motion from video. However, this approach is not causal since the receptive field is conditioned on past and future frames. Future prediction requires a causal structure to ensure that predictions do not depend on information from the future.

In this section, we first present an overview of the output 3D mesh representation. Then, we discuss the encoder model that learns a causal latent representation of human motion and an autoregressive model for future prediction in this latent space.
Lastly, we explain our training procedures.

\subsection{3D Mesh Representation}

We represent the 3D mesh with 82 parameters $\Theta = [\theta, \beta]$ consisting of pose and shape. We employ the SMPL 3D mesh body model \cite{SMPL}, which is a differentiable function $\mathcal{M}(\beta, \theta)\in \mathbb{R}^{6890\times 3}$ that outputs a triangular mesh with 6890 vertices given pose $\theta$ and shape $\beta$.
The pose parameters $\theta \in \mathbb{R}^{72}$ contain the global rotation of the body and relative rotations of 23 joints in axis-angle representation.
The shape parameters $\beta \in \mathbb{R}^{10}$ are the linear coefficients of a PCA shape space.
The SMPL function shapes a template mesh conditioned on $\theta$ and $\beta$, applies forward kinematics to articulate the mesh according to $\theta$, and deforms the surface via linear blend skinning. 
More details can be found in \cite{SMPL}.

We use a weak-perspective camera model $\Pi = [s, t_x, t_y]$ that represents scale and translation.
From the mesh, we can extract the 3D coordinates of $j$ joints $X\in \mathbb{R}^{j\times 3}= W\mathcal{M}(\beta, \theta)$ using a pre-trained linear regressor $W$. From the 3D joints and camera parameters, we can compute the 2D projection which we denote as $x \in \mathbb{R}^{j\times 2} = \Pi(X(\beta, \theta))$.

In this work, we use the SMPL mesh as a design decision, but many of the core concepts proposed could be extended to a skeletal model.

\subsection{Causal Model of Human Motion}
\label{sec:latent_representation}

\begin{table}[t]
    \resizebox{\columnwidth}{!}{
        
        \begin{tabular}{cccc}
            \toprule
            & \multicolumn{2}{c}{H36M} & Penn Action\\
            \cmidrule(lr){2-3} \cmidrule(lr){4-4}
            & MPJPE $\downarrow$  & \multicolumn{1}{c}{Reconst. $\downarrow$} & \multicolumn{1}{c}{PCK $\uparrow$} \\
            \midrule
            Causal model & 83.9 &	\textbf{56.7} &\textbf{	80.6} \\
            Kanazawa \etal \cite{kanazawa19hmmr} & \textbf{83.7}	& 56.9 &  79.6\\ 
            \bottomrule
        \end{tabular}

    }
    \caption{{{\bf Comparison of our causal temporal encoder with a non-causal model.} Although conditioned only on past context, our causal model performs comparably with a non-causal model that can see the past and the future. Both models have a receptive field of 13 frames, but our model uses edge padding for the convolutions while \cite{kanazawa19hmmr} use zero padding. MPJPE and Reconstruction error are measured in \textit{mm}. PCK \cite{yang2013pck} is a percentage.}}
      \vspace{-3mm}
    \label{tab:causal}
\end{table}

In this work, we train a neural autoregressive prediction model on the latent representation of 3D human motion encoded from the video. This allows seamless transition between conditioning on the past images frames and conditioning on previously generated future predictions. 

In order to learn the latent representation, we follow-up on our previous work \cite{kanazawa19hmmr}, which learns a latent encoding of 3D human motion from the temporal context of image sequences. That work uses a series of 1D convolutional layers over the time dimension of per-frame image features to learn an encoding of an image context, whose context length is equal to the size of the receptive field of the convolutions. However, because the goal was to simply obtain a smooth, temporally consistent representation of humans, that convolution kernel was centered, incorporating both past and future context. Since the goal in this work is to perform future prediction, we require our encoders to be causal, where the encoding of past temporal context at time $t$ is convolved only with elements from time $t$ and earlier in the previous layers \cite{bai2018empirical}. Here we discuss our causal encoder for human motion from video.

Our input is a video sequence of a person $V=\{I_t\}_{i=t}^{t-N}$ where each frame is cropped and centered around the subject. Each video sequence is paired with 3D pose and shape parameters or 2D keypoint annotations. We use a pre-trained per-frame feature extractor to encode each image frame $I_t$ into a feature vector $\phi_t$. To encode 3D human dynamics, we train a causal temporal encoder  $\f{movie}: \{\phi_{t-r}, \ldots, \phi_{t-1}, \phi_t \} \mapsto \Phi_t$. Intuitively, $\Phi_t$ represents a ``movie strip" that captures the temporal context of 3D human motion leading up to time $t$. This differs from the temporal encoder in \cite{kanazawa19hmmr} since the encoder there captures the context centered at $t$. 
Now that we have a representation capturing the motion up to time $t$, we train a 3D regressor
$\f{3D}: \Phi_t \mapsto \Theta_t$ that predicts the 3D human mesh at $t$ as well as the camera parameters $\Pi_t$.
The temporal encoder and the 3D regressors are trained with 3D losses on videos that have ground truth 3D annotations:
$\loss{3D} = \norm{X_t - \hat{X}_t}_2^2 + \norm{\theta_t - \hat{\theta}_t}_2^2 + \norm{\beta - \hat{\beta}}_2^2$.

However, datasets with 3D annotations are generally limited. 3D supervision is costly to obtain, requiring expensive instrumentation that often confines the videos captured to controlled environments that do not accurately reflect the complexity of human appearances in the real world. 
To make use of in-the-wild datasets that only have 2D ground truth or pseudo-ground truth pose annotations, we train our models with 2D re-projection loss \cite{3dinterpreter} on visible 2D keypoints: $\loss{2D} = \norm{v_t\odot(x_t - \hat{x_t})}_2^2$,
where $v_t\in \mathbb{R}^{j\times 2}$ is the visibility indicator over ground truth keypoints. We also use the factorized adversarial prior loss $\loss{adv prior}$ proposed in \cite{kanazawa18hmr, kanazawa19hmmr} to constrain the predicted poses to lie in the manifold of possible human poses.
We regularize our shape using the shape prior $\loss{beta loss}$ \cite{SMPLify}. Thus, for each frame $t$, the total loss is $L_t = \loss{3D} + \loss{2D} + \loss{adv prior} + \loss{beta loss}$.
As in \cite{kanazawa19hmmr}, we include a loss to encourage the model to predict a consistent shape:
$\loss{const} = \sum_{t=1}^{T - 1} \norm{\beta_t - \beta_{t+1}}_2^2$
and predict the mesh of nearby frames, encouraging the model to pay more attention to the temporal information in the movie strip at hand. With a receptive field of 13, \cite{kanazawa19hmmr} uses neighboring frames $\Delta t \in \{-5, +5\}$ whereas we use $\Delta t \in \{-10, -5\}$ since our model is causal.
Altogether, the objective function per sequence for the causal temporal encoder is
\begin{equation}
    \loss{movie} = \sum_t L_t  + \loss{const} + \sum_{\Delta t} L_{t + \Delta t}
\end{equation}

Comparison of our causal temporal encoder with \cite{kanazawa19hmmr} is in Table \ref{tab:causal}. Our causal model performs comparably despite not having access to future frames.

\subsection{Autoregressive prediction}
\label{sec:ar}
Now that we have a latent representation $\Phi_t$ of the motion leading up to a moment in time $t$, we wish to learn a prediction model that generates the future 3D human mesh model given the latent movie-strip representation of the input video $\Phi = \{\Phi_{i}, \Phi_{i-1}, ... \Phi_{i-r}\}$.   
We treat this problem as a sequence modeling task, where we model the joint distribution of the future latent representation as:
\begin{equation}
p(\Phi) = p(\Phi_{1}, \Phi_2, \dots, \Phi_{T}).
\end{equation}

One way of modeling the future distribution of 3D human motion $\Phi$ is as a product of conditional probabilities of its past:
\begin{equation}
p(\Phi) = \prod_{t=1}^T p(\Phi_t | \Phi_{1}, \dots, \Phi_{t-1}).
    \label{eq:model}
\end{equation}
In particular, following the recent success of temporal convolutional networks \cite{bai2018empirical,oord2016wavenet}, we also formulate this with 1D causal convolutional layers:
\begin{equation}
\tilde{\Phi}_t = \f{AR}(\Phi_{1}, \dots, \Phi_{t-1})
\end{equation}

In practice, we condition on the $r + 1$ past image features, where $r + 1$ is the receptive field size of the causal convolution. 
Since the future is available, this can be trained in a self-supervised manner via a distillation loss that encourages the predicted movie strips to be close to the real movie strips:
\begin{equation}
\loss{movie strip} = \norm{\Phi_t - \tilde{\Phi}_t},
\end{equation}
where $\Phi_t$ is the ground truth movie strip produced by the temporal encoder $\f{movie}$.
Moreover, since this latent representation should accurately capture the future state of the person, we use $\f{3D}$ to read out the predicted meshes from the predicted movie strips. Without seeing the actual images, $\f{AR}$ and $\f{3D}$ are unable to predict any meaningful camera predictions. To compute the re-projection loss without a predicted camera, we solve for the optimal camera parameters that best align the orthographically projected joints with the visible ground truth 2D joints $x_\text{gt}$:
\begin{equation}
\Pi^* = \underset{s, t_x, t_y}{\arg \min}\left \lVert \left(sx_\text{orth} + \begin{bsmallmatrix} t_x \\t_y \end{bsmallmatrix}\right) -x_\text{gt} \right \rVert_2^2,
\end{equation}
where $x_\text{orth}$ is the orthographically projected 3D joints ($X$ with the depth dimension dropped).

Now, we can apply all the losses from Section \ref{sec:latent_representation} to future prediction. 
In summary the total loss is:
\begin{equation}
    \loss{AR} = \sum_t L_t + \loss{const} + \sum_t \loss{movie strip}
\end{equation}

To better compare with methods that perform 3D prediction from 3D input, we also study a variant of our approach that makes autoregressive predictions directly in the pose space $\Theta$:
\begin{equation}
    \tilde{\Theta}_t = \f{AR}^\Theta(\Theta_{1}, \dots, \Theta_{t-1}).
\end{equation}

\begin{figure*}[t]
	\centering
	\includegraphics[width=1\textwidth]{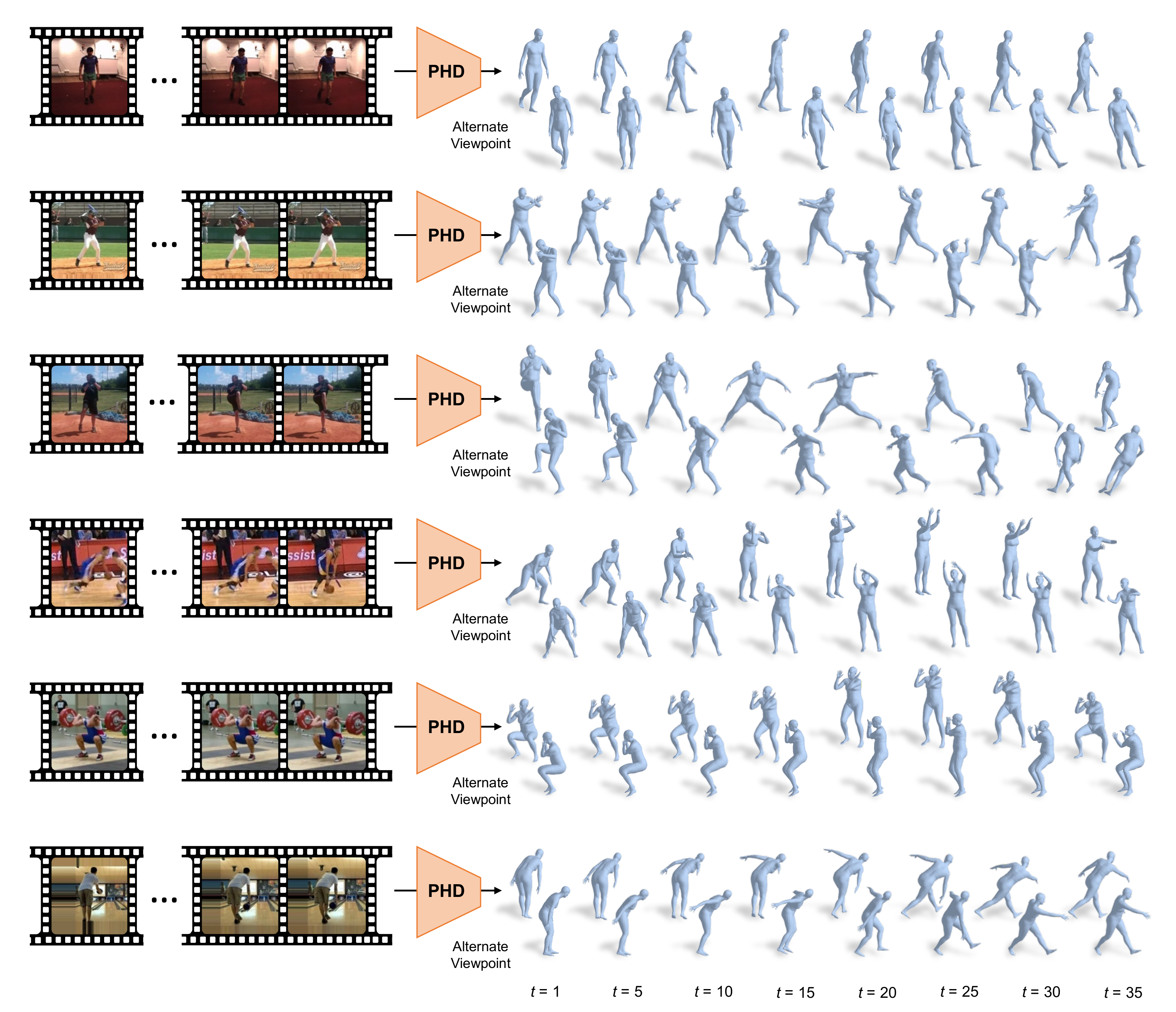}
	\vspace{-5mm}
	\caption{{{\bf Qualitative results of our approach on sequences from Human3.6M, Penn Action, and NBA.} In each sequence, the model takes as input 13 video frames (0.5 seconds) and  predicts the 3D motion in an autoregressive manner in the latent space.  The model is trained to predict for 25 time steps, but we show predictions made over 35 time steps from two different viewpoints. Results are sub-sampled at every 5 steps for space considerations. Please see the Supplementary Material for more results including the entire video sequences. We are able to capture periodic motions such as walking as well as complex sport motions including batting, pitching, shooting basketball hoops, performing a clean, and bowling.} }
	\vspace{-3mm}
	\label{fig:bigfig}

\end{figure*}

\subsection{Training Procedures}
We employed a two-step training strategy. We first trained the temporal encoder $\f{movie}$ and the 3D regressor $\f{3D}$, and then trained the autoregressive predictor $\f{AR}$. We froze the weights of the pre-trained ResNet from \cite{kanazawa18hmr} and trained the temporal encoder $\f{movie}$ and the 3D regressor $\f{3D}$ jointly on the task of estimating 3D human meshes from video. After training converged, we froze $\f{3D}$ and trained $\f{movie}$ and the autoregressive predictor $\f{AR}$ jointly.

To train the autoregressive model, we employ a curriculum-based approach \cite{AfourasCZ18a}. When training sequence generation models, it is common to use teacher forcing, in which the ground truth is fed into the network as input at each step. However, at test time, the ground truth inputs are unavailable, resulting in drifting since the model wasn't trained with its own predictions as input. To help address this, we train consecutive steps with the model's own output at previous time steps as inputs, similar to what is done in \cite{li2018auto}.
We slowly increase the number of consecutive predicted outputs fed as input to the autoregressive model, starting at 1 step and eventually hitting 25 steps.

While our approach can be conditioned on a larger past context by using dilated convolutions \cite{bai2018empirical}, our setting is bottlenecked by the length of the training videos. Recovering long human tracks from video is challenging due to occlusion and movement out of frame, and existing datasets of humans in the wild that we can train on tend to be short. Penn Action \cite{Zhang2013Penn} videos, for instance, have a median length of 56 frames. Since both the temporal encoder and autoregressor have a receptive field of 13, 25 was near the upper-bound of the number of autoregressive predictions we could make given our data. See the supplemental materials for further discussion.

%% file: phd_4_eval.tex
\section{Evaluation}
\label{sec:evaluation}

\begin{table*}[ht]
 \begin{minipage}[t]{0.71\linewidth}
 	\vspace{0pt}
	\centering
	\resizebox{\textwidth}{!}{
	\begin{tabular}{ccccccccccc}
		\toprule
		& \multicolumn{5}{c}{Human3.6M \tabspace Reconst. $\downarrow$} & \multicolumn{5}{c}{Penn Action \tabspace PCK $\uparrow$}  \\
		\cmidrule(lr){2-6} \cmidrule(lr){7-11} 
		Method & 1 & 5 & 10 & 20 & 30 & 1 & 5 & 10 & 20 & 30 \\
		\midrule
		AR on $\Phi$ & 57.7 & 59.5 & \textbf{61.1} & 62.1 & \textbf{65.1} & \textbf{81.2} & \textbf{80.0} & \textbf{79.0} & \textbf{78.2} & \textbf{77.2}\\
		No $\loss{movie strip}$ & \textbf{56.9} & \textbf{59.2} & \textbf{61.1} & \textbf{61.9} & 65.3 & 80.4 & 78.7 & 77.6 & 76.8 & 75.6\\ 
        AR on $\Theta$ & 57.8 & 	61.7 & 	66.7 & 	75.3 & 	82.6 & 	79.9 & 	74.3 & 	68.5 & 61.4 & 	56.5\\ 
        Constant & 59.7 & 	65.3 & 	72.8 & 	84.3 & 	90.4 & 	78.3 & 	71.7 & 	64.9 & 	56.2 & 	49.7\\ 
        NN & 90.3 &	95.1 & 100.6 & 108.6 & 	114.2 & 	63.2 & 	61.5 & 	60.6 & 	58.7 & 	57.8\\
        \bottomrule
	\end{tabular}
	}
	\caption{\small{{\bf Comparison of autoregressive predictions and various baselines (with Dynamic Time Warping).} We evaluate our model with autoregressive prediction in the movie strip latent space $\Phi$ (AR on $\Phi$), an ablation in the latent space without the distillation loss (No $\loss{movie strip}$), and predictions in the pose space $\Theta$ (AR on $\Theta$). We also compare with the no-motion baseline (Constant) and Nearest Neighbors (NN).}}
	\label{tab:omega_baseline}
\end{minipage}
\hspace{0.7mm}
\begin{minipage}[t]{0.27\linewidth}
	\vspace{0pt}
    \centering
    \resizebox{\textwidth}{!}{
    \begin{tabular}{ccc}
        \toprule
         & H3.6M $\downarrow$ & Penn $\uparrow$ \\
        \midrule
        AR on $\Phi$ &  61.2 & 77.2\\
        AR on $\Theta$ & 65.9 & 67.8 \\
        \cite{chao2017forecasting} & - & 68.1\\
        \cite{kanazawa19hmmr} & 65.3 & 67.8\\
        \bottomrule
    \end{tabular}
    }
    \captionof{table}{\small{\textbf{Comparison with single frame future prediction.} We compare our method with the future prediction from single image proposed in Chao \etal \cite{chao2017forecasting} and Kanazawa \etal \cite{kanazawa19hmmr}. All methods are evaluated 5 frames into the future without DTW.}}
    \label{tab:hmmr_comparison}
\end{minipage}
\end{table*}
\begin{figure*}[th]
	\centering
	\includegraphics[width=\textwidth]{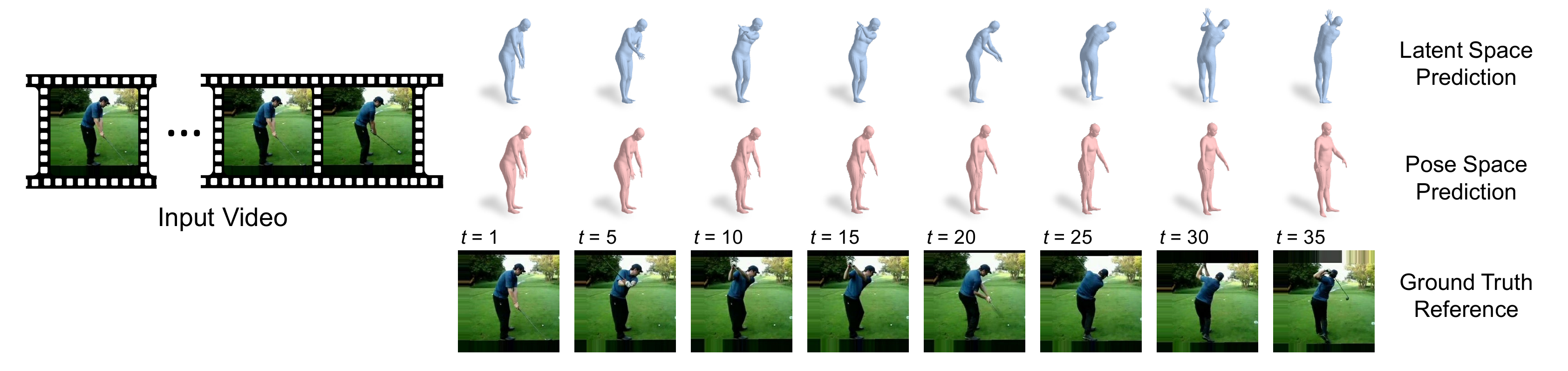}
	\vspace{-5mm}
	\caption{{{\bf Comparison between predictions in latent versus pose space.} The blue meshes are predictions made in the latent movie strip space, while the pink meshes are predictions made directly in the pose space. We observe that while the latent space prediction does not always predict the correct tempo, it generally predicts the correct sequence of poses.}}
	\vspace{-3mm}
	\label{fig:omega_comp}
\end{figure*}

\begin{figure*}[th]
	\begin{subfigure}[t]{0.21\textwidth}
		\centering
		\includegraphics[width=0.7\textwidth]{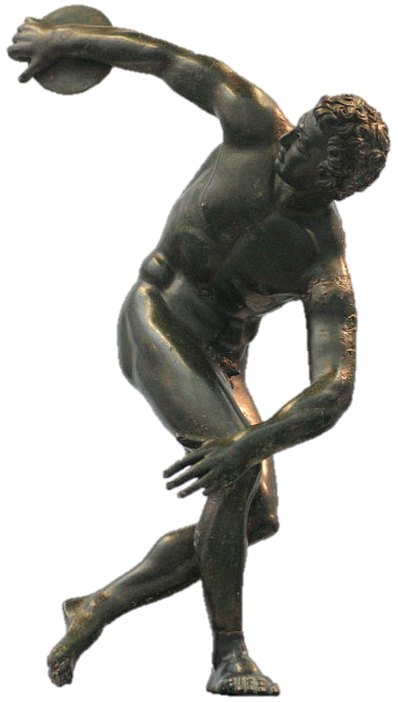}
		\vspace{-1mm}
		\caption{Bronze reproduction of Myron's \textit{Discobolus}.\\\footnotesize{\textbf{Source:}  Matthias Kabel}}
		\label{fig:discobolus}	
	\end{subfigure}%
	\hfill
	\begin{subfigure}[t]{0.75\textwidth}
	    \centering
		\includegraphics[width=1\textwidth]{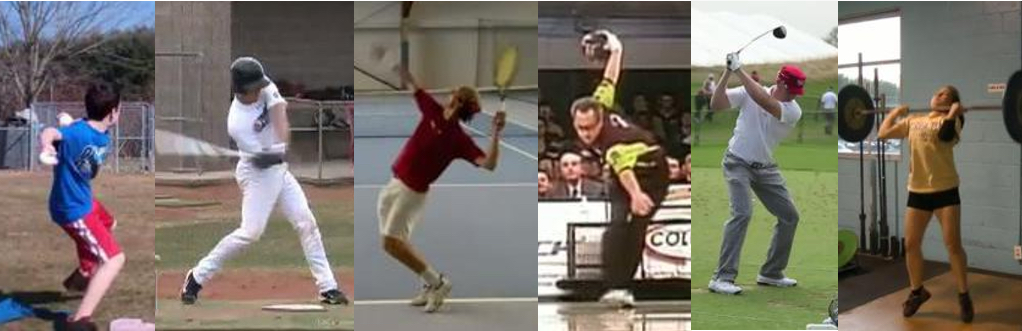}
		\caption{Penn Action frames with largest improvement in accuracy.}
		\label{fig:penn_statues}
	\end{subfigure}
	\vspace{-3mm}
	\caption{\textbf{Discovered ``Statue" moments.} Classical sculptures of athletes often depict the transient moment when the action becomes immediately clear to the viewer (\ref{fig:discobolus}). The frame that gives rise to the largest improvement in prediction accuracy exhibits similar qualities to these statues (\ref{fig:penn_statues}). Here, we visualize the first frame after the best input conditioning for several sequences.}
	\label{fig:statues}
	\vspace{-3mm}
\end{figure*}

In this section, we present quantitative and qualitative results of 3D mesh motion generation from video.

\subsection{Experimental Setup}
\noindent\textbf{Network Architecture.}
We use the pre-trained ResNet-50  provided by \cite{kanazawa18hmr} as our image feature encoder and use the average pooled feature of the last layer as our $\phi \in \mathbb{R}^{2048}$.
The causal  temporal encoder $\f{movie}$ and the autoregressive predictor $\f{AR}$ both have the same architecture, consisting of 3 residual blocks. Following \cite{he2016resnet}, each block consists of GroupNorm, ReLU, 1D Convolution, GroupNorm, ReLU, 1D Convolution. Each 1D Convolution uses a kernel size of 3 and a filter size of 2048.
Unlike \cite{kanazawa19hmmr} which use zero padding, we use edge padding for the 1D convolutions.
In total, the 3 residual blocks induce a receptive field of 13 frames (about 0.5 seconds at 25 fps). To make the encoder causal, we shift the output indices so that the prediction for time $t$ corresponds to the output that depends only on inputs up to $t$ from previous layers. The movie-strip representation $\Phi$ also has 2048 dimensions. The autoregressive variant model that predicts the future in the 3D mesh space $\f{AR}^\Theta$ has the same architecture except it directly outputs the 82D mesh parameters $\Theta$. For the 3D regressor $\f{3D}$, we use the architecture in \cite{kanazawa18hmr}.

\noindent\textbf{Datasets.} We train on 4 datasets with different levels of supervision. The only dataset that has ground truth 3D annotations is Human3.6M \cite{Human36m:2014}, which contains videos of actors performing various activities in a motion capture studio. We use Subjects 1, 6, 7, and 8 as the training set, Subject 5 as the validation set, and Subjects 9 and 11 as the test set. Penn Action \cite{Zhang2013Penn} and NBA \cite{kanazawa19hmmr} are datasets with 2D ground truth keypoint annotations of in-the-wild sports videos. Penn Action consists of 15 sports activities such as golfing or bowling, while NBA consists of videos of professional basketball players attempting 3-points  shots. InstaVariety \cite{kanazawa19hmmr} is a large-scale dataset of internet videos scraped from Instagram with \textit{pseudo-}ground truth 2D keypoint annotations from OpenPose \cite{cao2017realtime}. We evaluate on the Human3.6M and Penn Action datasets which have 3D and 2D ground truth respectively.
We train on all of these videos together in an action- and dataset-agnostic manner.

\subsection{Quantitative Evaluation}

\noindent\textbf{Dynamic Time Warping.} 
Predicting the future motion is a highly challenging task. Even if we predict the correct type of motion, the actual start time and velocity of the motion are still ambiguous. Thus, for evaluation we employ Dynamic Time Warping (DTW), which is often used to compute the similarity between sequences that have different speeds. In particular, we compute the similarity between the ground truth and predicted future sequence after applying the optimal non-linear warping to both sequences. The optimal match maximizes the similarity of the time-warped ground truth joints and the time-warped predicted joints subject to the constraint that each set of ground truth joints must map to at least one set of predicted joints and vice-versa. In addition, the indices of the mapping must increase monotonically. For detailed evaluation without DTW as well as an example alignment after applying DTW, please see the Supplementary Materials.

\noindent\textbf{Evaluation Procedures and Metrics.}
For Human3.6M where ground truth 3D annotations exist, we report the reconstruction error in \textit{mm} by computing the mean per joint position error after applying Procrustes Alignment. For Penn Action which only has 2D ground truth annotations, we measure the percentage of correct keypoints (PCK) \cite{yang2013pck} at $\alpha = 0.05$. 
We begin making autoregressive predictions starting from every 25th frame for Human3.6M and starting from every frame for Penn Action after conditioning on 15 input frames. Although we train with future prediction up to 25 frames, we evaluate all metrics for poses through 30 frames into the future.

\noindent\textbf{Baselines.}
We propose a no-motion baseline (Constant) and a Nearest Neighbor baseline (NN), which we evaluate in Table \ref{tab:omega_baseline}. The no-motion baseline freezes the estimated pose corresponding to the last observed frame. We use the causal temporal encoder introduced in Section \ref{sec:latent_representation} for the 3D pose and 2D keypoint estimations.

The Nearest Neighbor baseline takes a window of input conditioning frames from the test set and computes the closest sequence in the training set using Euclidean distance of normalized 3D joints. The subsequent future frames are used as the prediction.
We estimate the normalized 3D joints (\ie mean SMPL shape) for each frame using our temporal encoder.
See the Supplementary Materials for examples of Nearest Neighbors predictions.

\noindent\textbf{Prediction Evaluations.}
In Table \ref{tab:omega_baseline}, we compare ablations of our method with both baselines. 
We evaluate our predictions in both the latent space and the pose space as proposed in Section \ref{sec:ar}. The results show that predictions in the latent space significantly outperform the predictions in the pose space, with the difference becoming increasingly apparent further into the future. This is unsurprising since the pose can always be read from the latent space, but the latent space can also capture additional information such as image context that may be useful for determining the action type. Thus, performance in the latent space should be at least as good as that in the pose space.

We also evaluate the effect of the distillation loss by removing $\loss{movie strip}$. The performance diminishes slightly on Penn Action but is negligibly different on Human3.6M. It is possible that the latent representation learned by $\f{movie}$ is more useful in the absence of 3D ground truth.

Finally, our method in the latent space significantly outperforms both baselines. The no-motion baseline performs reasonably at first since it's initialized from the correct pose but quickly deteriorates as the frozen pose no longer matches the motion of the sequence. On the flip side, the Nearest Neighbors baseline performs poorly at first due to the difficulty of aligning the global orientation of the root joint. However, on Penn Action, NN often identifies the correct action and eventually outperforms the no-motion baseline and auto-regressive predictions in the pose space.

\noindent\textbf{Comparison with Single Frame future prediction.} In Table  \ref{tab:hmmr_comparison}, we compare our method with the single frame future prediction in \cite{chao2017forecasting} and \cite{kanazawa19hmmr}. To remain comparable, we retrained \cite{chao2017forecasting} to forecast the pose 5 frames into the future and evaluate all methods on the 5th frame past the last observed frame. Note that our method is conditioned on a sequence of images in an auto-regressive manner while \cite{kanazawa19hmmr} and \cite{chao2017forecasting} hallucinate the future 3D pose and 2D keypoints respectively from a single image. Our method produces significant gains on the Penn Action dataset where past context is valuable for future prediction on fast-moving sports sequences.

\subsection{Qualitative Evaluation}

\noindent\textbf{Qualitative Analysis.}
We show qualitative results of our proposed method in the latent space on videos from Human3.6M, Penn Action, and NBA in Figure \ref{fig:bigfig}.
We observe that the latent space model does not always regress the correct tempo but usually predicts the correct type of motion. On the other hand, the pose space model has significant difficulty predicting the type of motion itself unless the action is obvious from the pose (\eg Situps and Pushups). See Figure \ref{fig:omega_comp} and the supplementary for comparisons between the latent and pose space models.

Discussion of failure modes can be found in the supplementary. More video results are available on our project webpage\footnote{Project Webpage: \href{https://jasonyzhang.com/phd/}{jasonyzhang.com/phd}} and at \href{https://youtu.be/aBKZ7HMF1SM}{youtu.be/aBKZ7HMF1SM}.

\noindent\textbf{Capturing the ``Statue" Moment.}
Classical sculptures are often characterized by dynamic poses ready to burst into action. Myron's \textit{Discobolus} (Figure \ref{fig:discobolus}) is a canonical example of this, capturing the split-second before the athlete throws the discus \cite{jenkins2012discobolus}. 
We show that the proposed framework to predict 3D human motion from video can be used to discover such Classical ``statue" moments from video, by finding the frame that spikes the prediction accuracy. In Figure \ref{fig:statues}, we visualize frames from Penn Action when the prediction accuracy increases the most for each sequence. Specifically, for each conditioning window for every sequence in Penn Action, we computed the raw average future prediction accuracy for the following 15 frames.
Then, we computed the per-frame change in accuracy using a low-pass difference filter and selected the window with the largest improvement. We find that the frame corresponding to the timestep when the accuracy improves the most effectively captures the ``suggestive" moments in an action.

%% file: phd_5_conclusion.tex
\section{Conclusion}
\label{sec:conclusion}
In this paper, we presented a new approach for predicting 3D human mesh motion from a video input of a person.
We train an autoregressive model on the latent representation of the video, which allows the input conditioning to transition seamlessly from past video input  to previously predicted futures. In principle, the proposed approach could predict arbitrarily long sequences in an autoregressive manner using 3D or 2D supervision. 
Our approach can be trained on motion capture video in addition to in-the-wild video with only 2D annotations.

Much more remains to be done. One of the biggest challenges is that of handling multimodality since there can be multiple possible futures. This could deal with inherent uncertainties such as speed or type of motion.
Other challenges include handling significant occlusions and incorporating the constraints imposed by the affordances of the 3D environment.

\noindent\textbf{Acknowledgments.}
We would like to thank Ke Li for insightful discussion and Allan Jabri and Ashish Kumar for valuable feedback. We thank Alexei A. Efros for the statues.
This work was supported in part by Intel/NSF VEC award IIS-1539099 and BAIR sponsors. 

%% file: phd_6_appendix.tex
\begin{figure*}[t]
	\centering
	\includegraphics[width=\textwidth]{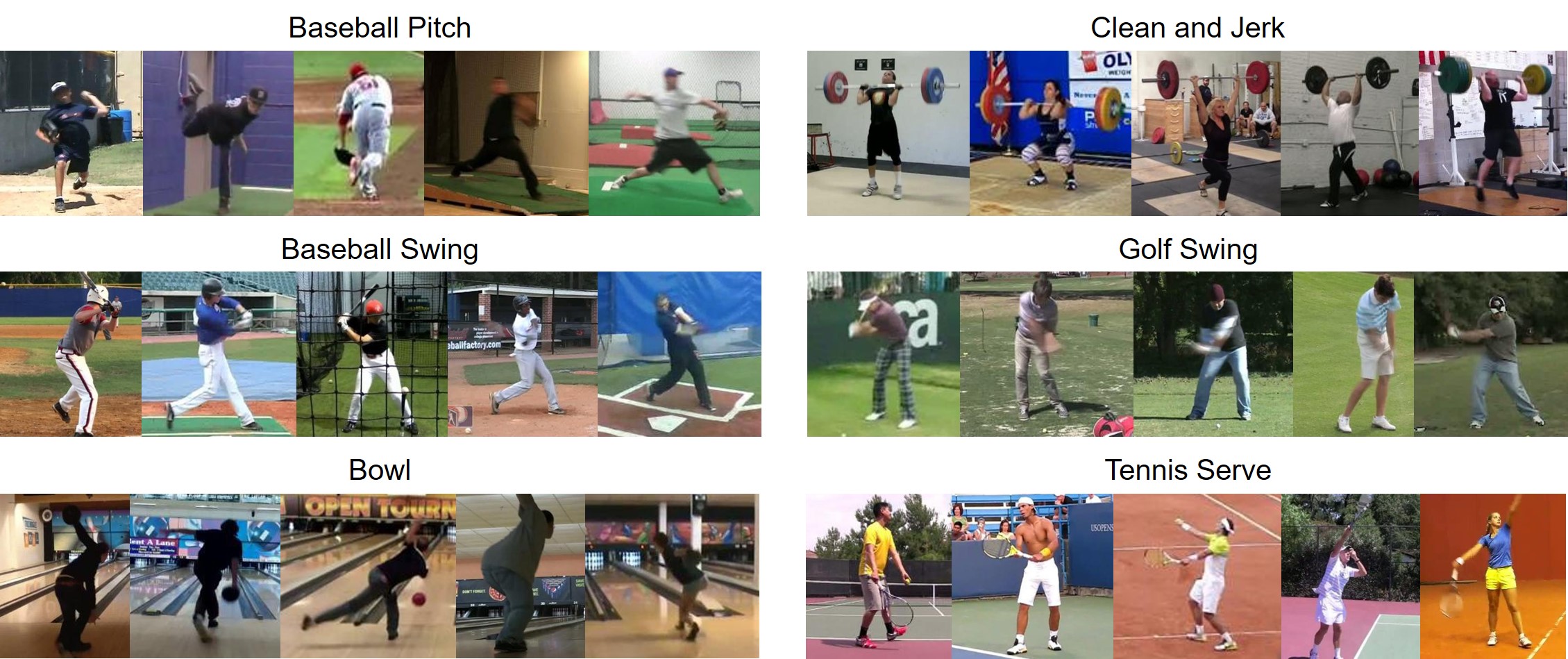}
	\caption{{{\bf Random Samples of Discovered ``Statues."} We show 5 \textit{random} samples of the statue discovery on 6 action categories from Penn Action that have fast, acyclic motion. For each sequence, we discovered the statue pose by finding the conditioning window when the prediction accuracy improves the most. Here, we visualize the first frame after the best input conditioning.
	}}
	\vspace{-3mm}
	\label{fig:statues_random}
\end{figure*}
\begin{figure*}
	\vspace{-2mm}
	\centering
	\includegraphics[width=\textwidth]{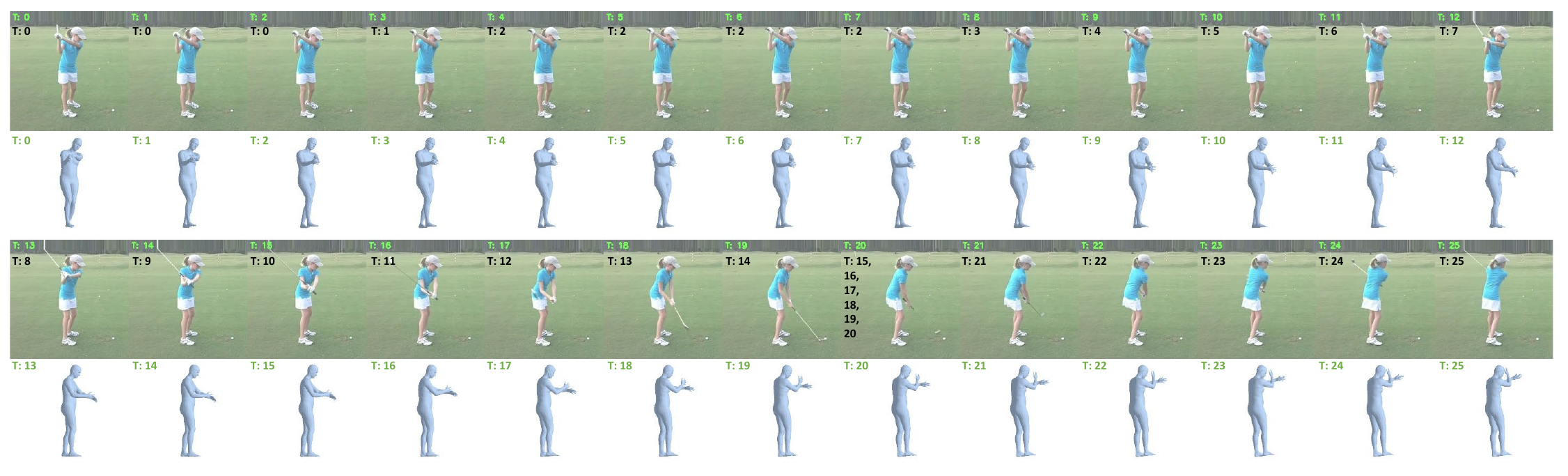}
	\vspace{-7mm}
	\caption{{{\bf Motion sequence after alignment via Dynamic Time Warping.} Dynamic Time Warping is often used to compute similarity between sequences that may have different speeds. Here, we show a \textit{Golf Swing} sequence in which the prediction model produces the correct action type but starts the swing too soon. The optimal match produced using Dynamic Time Warping reduces the penalty of mistiming the motion. Each frame of the ground truth sequence is matched with at least one frame of the predicted sequence and vice-versa. \textbf{Green:} Index of ground truth image and predicted mesh before applying Dynamic Time Warping. \textbf{Black:} Index of predicted mesh that corresponds to the ground truth image after applying Dynamic Time Warping.
	}}
	\label{fig:dtw}
	
\end{figure*}
\begin{figure*}[!b]
	\centering
	\includegraphics[width=0.4\textwidth]{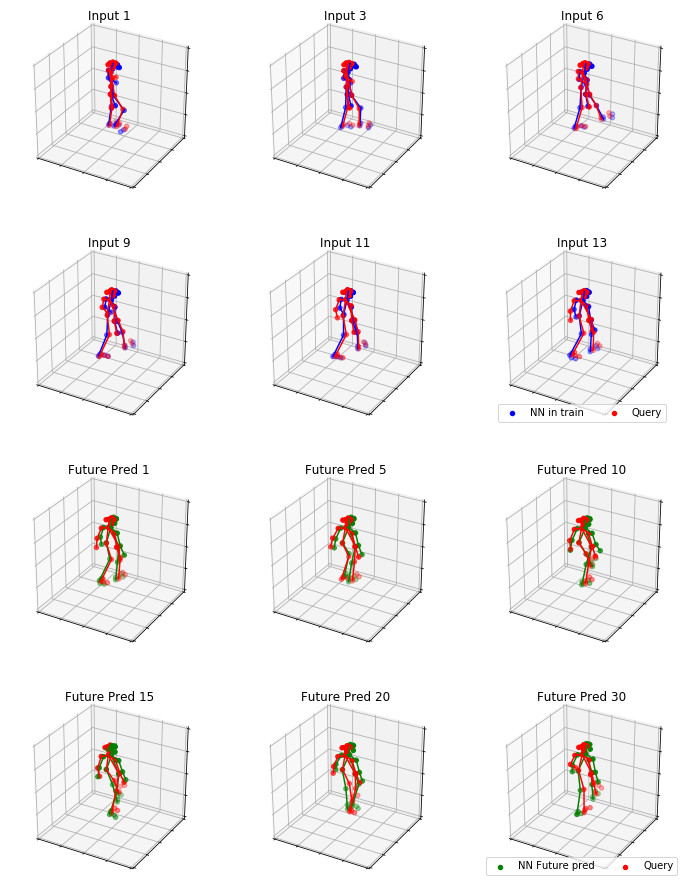}\hspace{1cm}
	\includegraphics[width=0.4\textwidth]{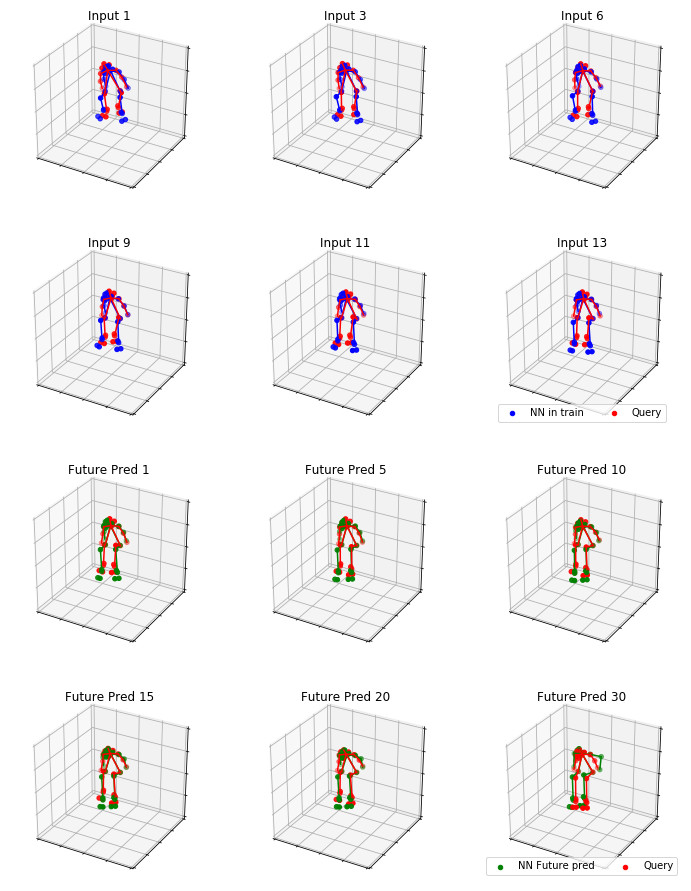}
	\caption{{{\bf Nearest Neighbor Future Predictions on Human3.6M.} For each sequence in the test set (red), we search for the best matching sequence in the training set (blue) and use the succeeding poses as the future prediction (green). \textbf{Left:} NN for a \textit{Walking} sequence. While the query has good fit, the NN prediction drifts further from the ground truth over time. \textbf{Right:} NN for a \textit{Sitting Down} sequence.
	}}
	\vspace{-3mm}
	\label{fig:nn_h36m}
\end{figure*}
\begin{figure*}[!b]
  \centering
  \includegraphics[width=0.4\textwidth]{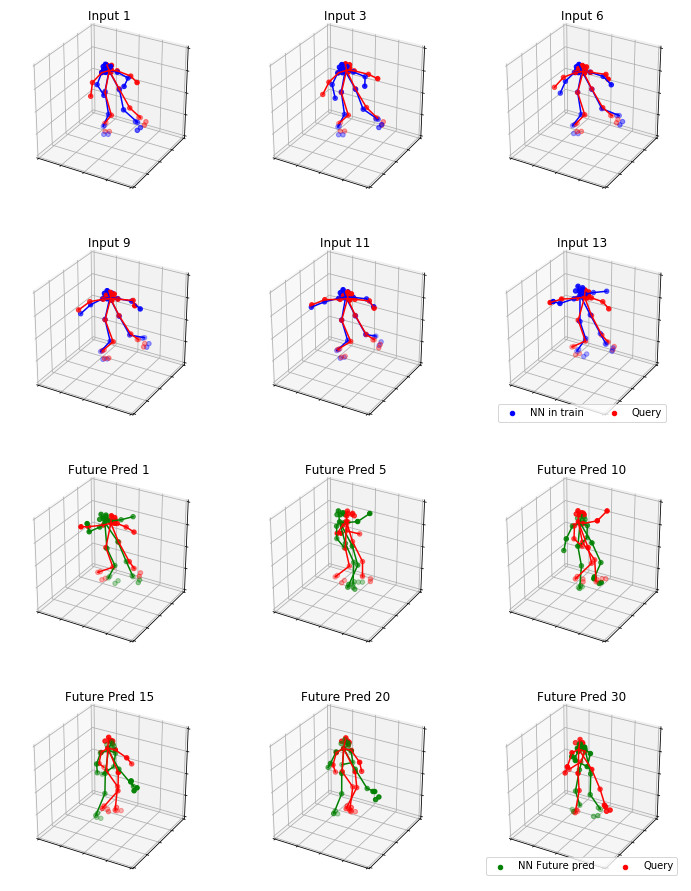}\hspace{1cm}
  \includegraphics[width=0.4\textwidth]{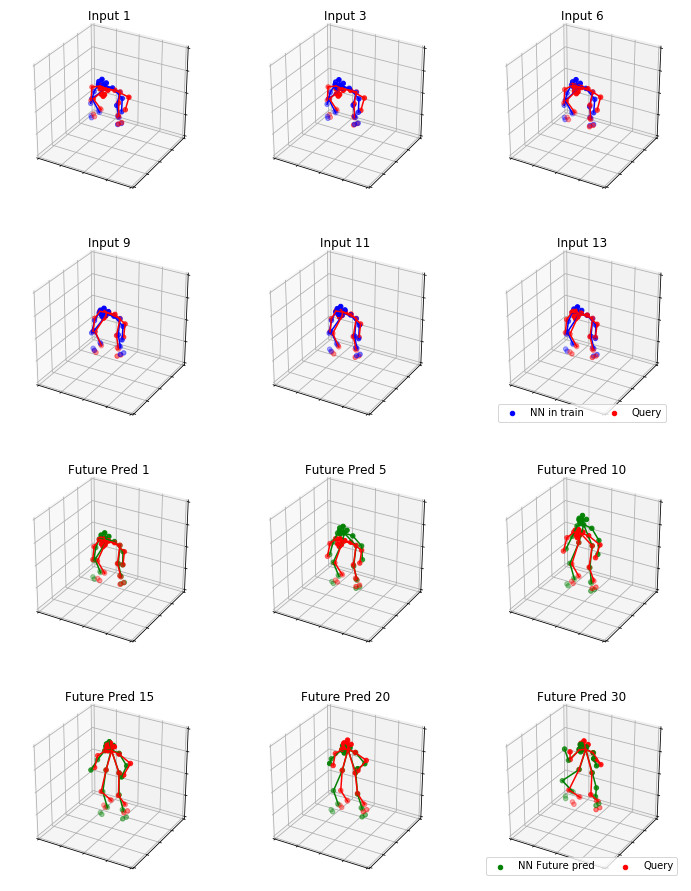}
  \caption{{{\bf Nearest Neighbor Future Predictions on Penn Action.} For each sequence in the test set (red), we search for the best matching sequence in the training set (blue) and use the succeeding poses as the future prediction (green). \textbf{Left:} NN for a \textit{Baseball Pitch} sequence. The predicted motion is faster than the ground truth motion. \textbf{Right:} NN for a \textit{Clean and Jerk}. The NN aligns well with the ground truth motion.
  }}
  \vspace{-3mm}
  \label{fig:nn_penn}
\end{figure*}
\begin{figure*}[ht]
  \centering
  \includegraphics[width=\textwidth]{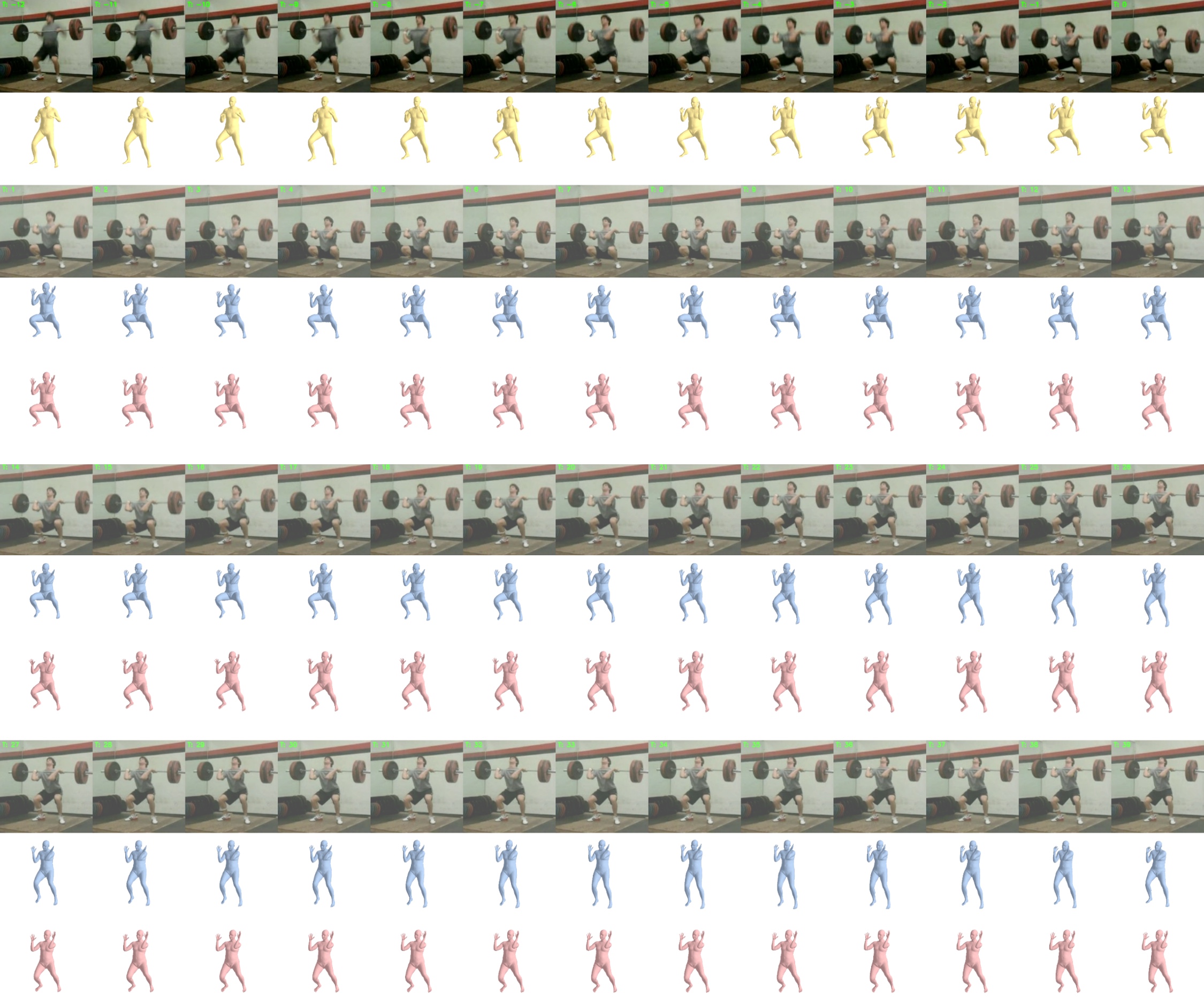}
  
  \caption{{{\bf Comparison of autoregressive models on performing a clean.} For simple motions, predictions in both the latent space and pose space perform reasonably. The first row of images shows the input sequence, and the rest of the images are ground truth for reference. We illustrate the conditioning with yellow meshes which are read out from the ground truth movie strips. The blue meshes show predictions in the latent space while the pink meshes show predictions in the pose space.
  }}
  \vspace{-3mm}
  \label{fig:clean_comparison}
\end{figure*}
\begin{figure*}[ht]
  \centering
  \includegraphics[width=\textwidth]{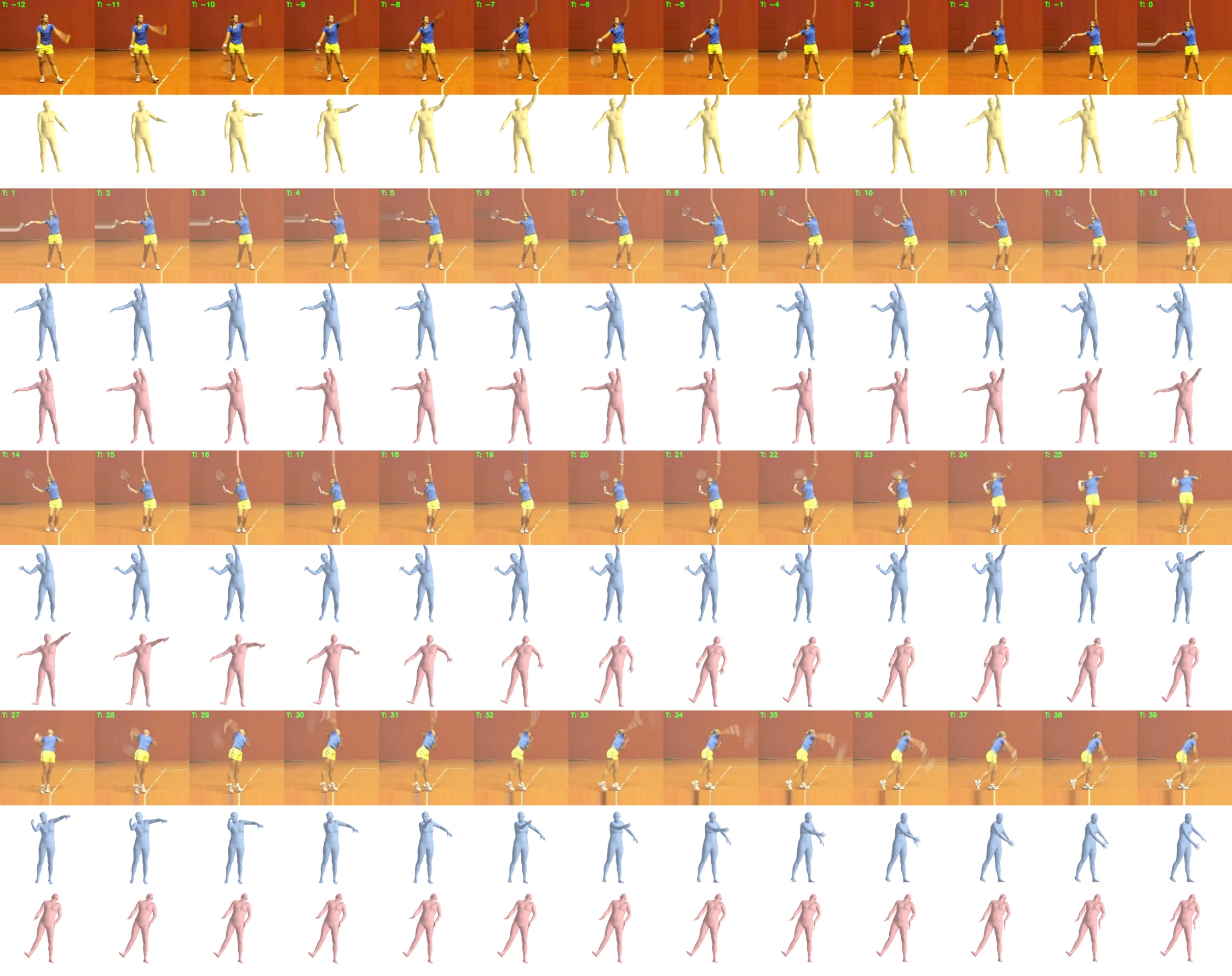}
  
  \caption{{{\bf Comparison of autoregressive models on a tennis serve.} For complex motions, predictions in latent space work reasonably well while predictions in the pose space struggle with identifying the action and motion. The first row of images shows the input sequence, and the rest of the images are ground truth for reference. We illustrate the conditioning with yellow meshes which are read out from the ground truth movie strips. The blue meshes show predictions in the latent space while the pink meshes show predictions in the pose space.
  }}
  \vspace{-3mm}
  \label{fig:serve_comparison}
\end{figure*}
\begin{figure*}[ht]
  \centering
    \includegraphics[width=\textwidth]{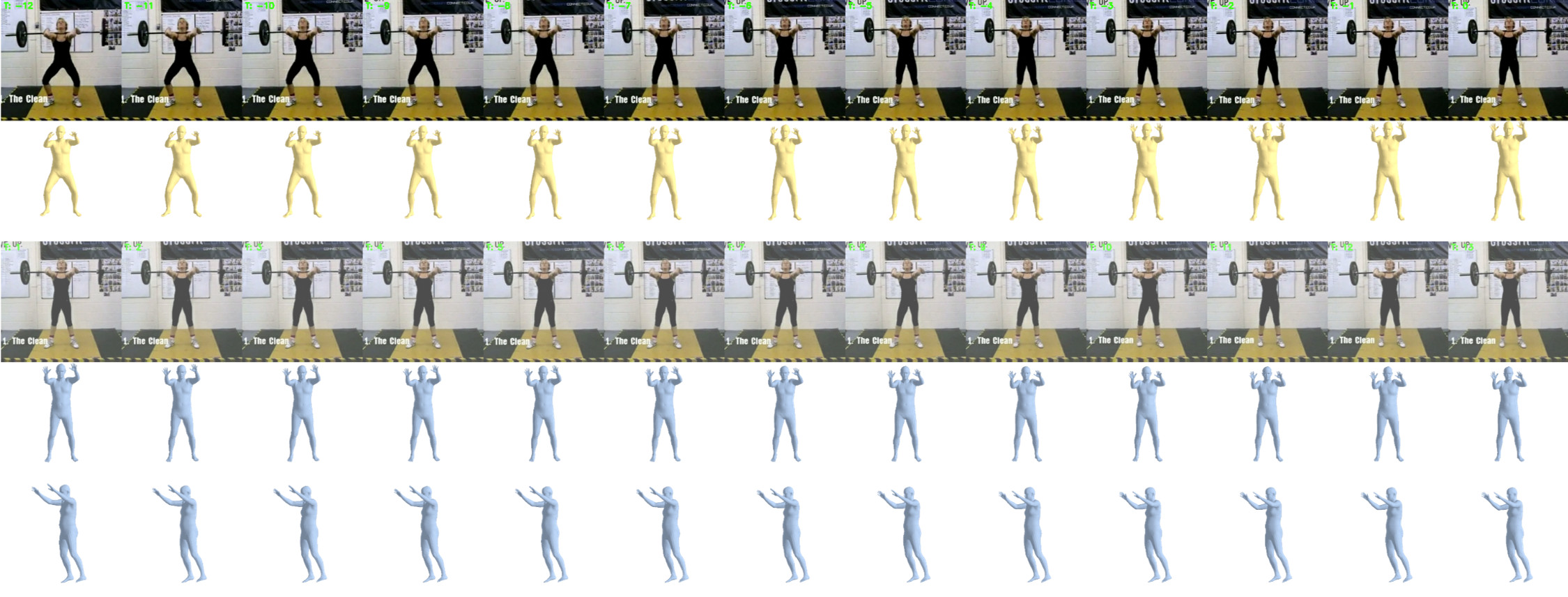}

  \caption{{{\bf Failure Mode: Ambiguity of 2D keypoints.}
  		In-the-wild data is generally labeled only with 2D keypoints, which can have multiple 3D interpretations. We rely on an adversarial prior to produce realistic poses. Here, our model predicts motion that incorrectly extends the subject's arms, but it is still anatomically plausible and projects to the correct 2D keypoints. The first row of images shows the input sequence, and the rest of the images are ground truth for reference. We illustrate the conditioning with yellow meshes which are read out from the ground truth movie strips. The blue meshes show predictions in the latent space from two different viewpoints.
  }}
  \vspace{-3mm}
  \label{fig:failure_ambiguous2d}
\end{figure*}
\begin{figure*}[ht]
  \centering
    \includegraphics[width=\textwidth]{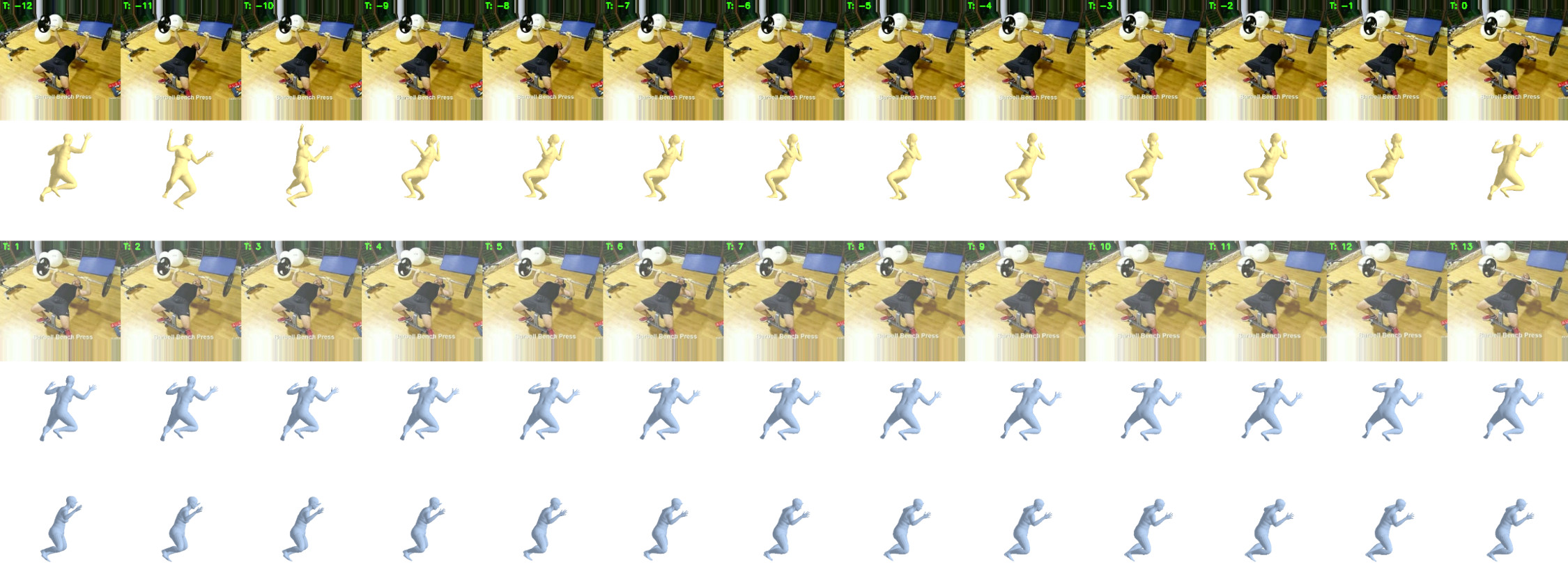}

  \caption{{{\bf Failure mode: Poor quality conditioning.} Our auto-regressive model is conditioned on the input movie strips from our temporal encoder. Mistakes made by the temporal encoder due to unusual viewpoints thus carry over to our prediction model. Here, the benching sequence is recorded from a top-down view, which is rarely encountered in the training data. The first row of images shows the input sequence, and the rest of the images are ground truth for reference. We illustrate the conditioning with yellow meshes which are read out from the ground truth movie strips. The blue meshes show predictions in the latent space from two different viewpoints.}}
  		
  \vspace{-3mm}
  \label{fig:failure_bad_cond}
\end{figure*}
\begin{figure*}[ht]
  \centering
    \includegraphics[width=\textwidth]{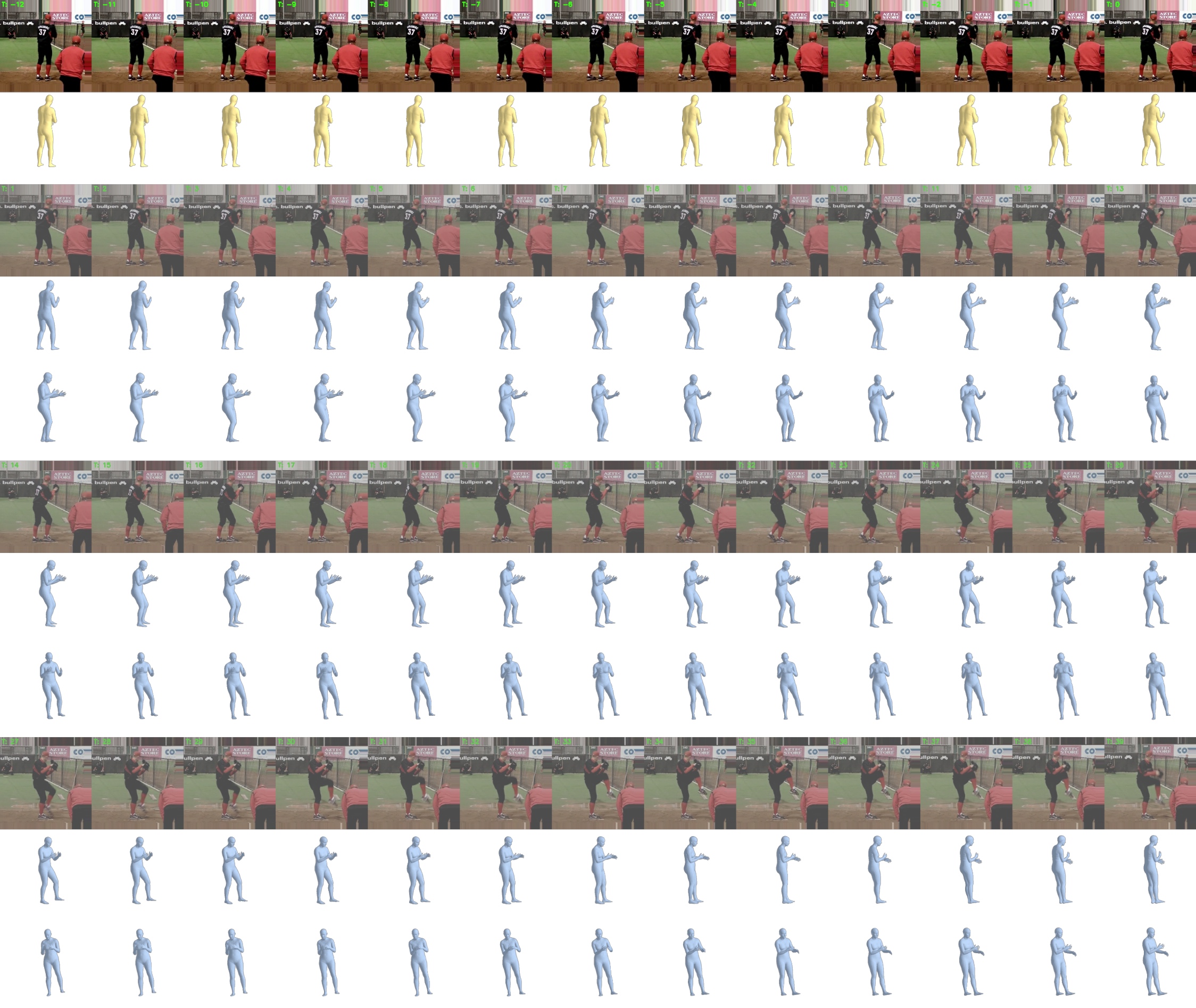}

  \caption{{{\bf Failure Mode: Conditioned on little motion.} Most sports actions in the Penn Action dataset begin with a short period with no motion as the player gets ready to pitch a ball, waits to bat, or prepares to golf. Thus, it is challenging to predict when the motion should begin when conditioned on frames corresponding to little motion. Here, the input frames show the pitcher barely moving, so our model predicts no motion while the athlete does begin to pitch later in the sequence. The first row of images shows the input sequence, and the rest of the images are ground truth for reference. We illustrate the conditioning with yellow meshes which are read out from the ground truth movie strips. The blue meshes show predictions in the latent space from two different viewpoints.
  }}
  \vspace{-3mm}
  \label{fig:failure_no_motion}
\end{figure*}
\begin{figure*}[ht]
	\centering
	\includegraphics[width=\textwidth]{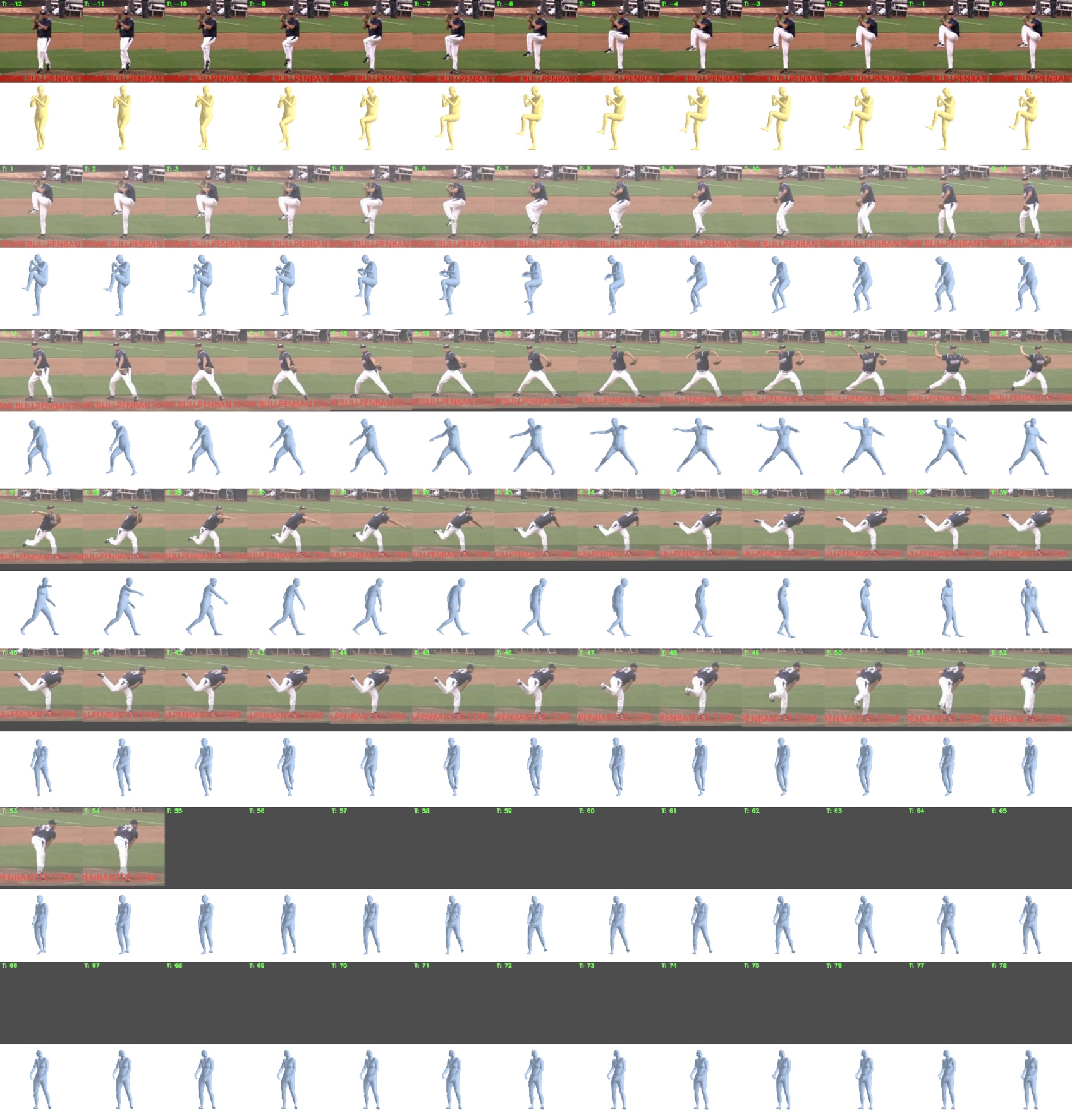}
	
	\caption{{{\bf Failure Mode: Drifting past 35 frames.} Due to the limited length of sequences in our training data, we train with future predictions up to 25 frames into the future. We observe that our model is capable of predicting outputs that look reasonable qualitatively until around 35 frames into the future. Training with longer sequences should alleviate this issue. The first row of images shows the input sequence, and the rest of the images are ground truth for reference. We illustrate the conditioning with yellow meshes which are read out from the ground truth movie strips. The blue meshes show predictions in the latent space.}}

	\vspace{-3mm}
	\label{fig:failure_drift}
\end{figure*}

In this section, we provide:
\begin{itemize}[noitemsep, nolistsep]
	\item Discussion of the implementation details with limited sequence length in Section \ref{sec:edge}.
	\item A random sample of discovered ``Statue" poses from Penn Action in Figure \ref{fig:statues_random}.
	\item An example of Dynamic Time Warping in Figure \ref{fig:dtw}.
	\item Per-action evaluation of Human3.6M (Table \ref{tab:h36m_sequences_dtw}) and Penn Action (Table \ref{tab:penn_sequences_dtw}) \textbf{with} Dynamic Time Warping.
	\item Per-action evaluation of Human3.6M (Table \ref{tab:h36m_sequences_nodtw}) and Penn Action (Table \ref{tab:penn_sequences_nodtw}) \textbf{without} Dynamic Time Warping.
	\item A comparison of our method with Constant and Nearest Neighbor baseline \textbf{without} Dynamic Time Warping in Table \ref{tab:all_baselines_nodtw}.
	\item A visualization of Nearest Neighbor Predictions in Human3.6m (Figure \ref{fig:nn_h36m}) and Penn Action (Figure \ref{fig:nn_penn}).
	\item A comparison of autoregressive predictions in the latent space versus pose space in Figures \ref{fig:clean_comparison} and \ref{fig:serve_comparison}.
	\item Discussion of failure modes such as ambiguity of 2D keypoints (Figure \ref{fig:failure_ambiguous2d}), poor conditioning (Figure \ref{fig:failure_bad_cond}), little motion in conditioning (Figure \ref{fig:failure_no_motion}), and drifting (Figure \ref{fig:failure_drift}).
\end{itemize}

\begin{figure*}[t]
	\centering
	\vspace{-3mm}
    \includegraphics[width=0.85\textwidth]{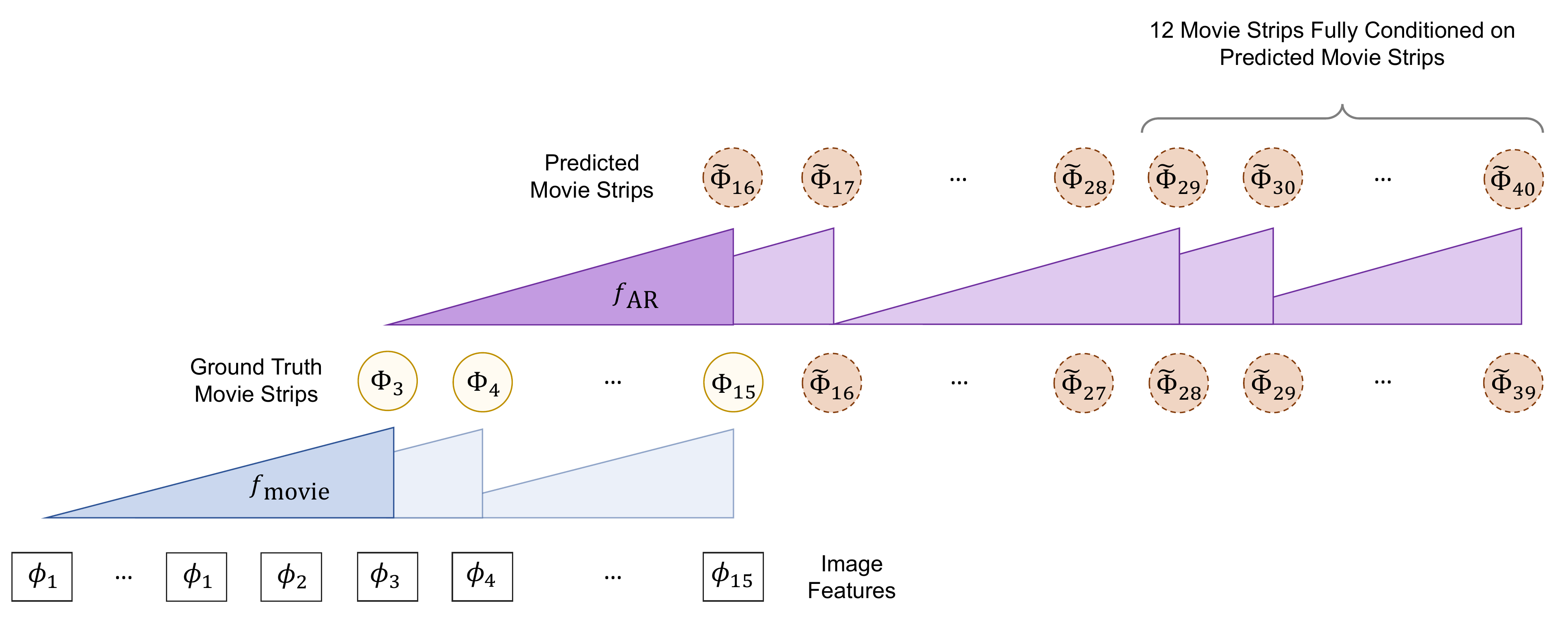}
    \vspace{-2mm}
	\caption{\textbf{Illustration of the full field of view of the proposed architecture.} 
	Since $f_\text{AR}$ and $f_\text{movie}$ each have a receptive field of 13, it is theoretically possible for $f_\text{AR}$ to be conditioned on 25 ground truth images. However, we train on videos that have a minimum length of 40 frames. In order to predict 25 frames into the future, we reduce the number of conditioned images to 15 by edge padding the first set of image features. See Section \ref{sec:edge} for more details.}
	\vspace{-3mm}
	\label{fig:conditoning_actual}
\end{figure*}

\subsection{Implementation Details of Sequence Length}
\label{sec:edge}

As discussed in the main paper, while our approach can be conditioned on a larger past context by using dilated convolutions, our setting is bottlenecked by the length of the training videos. Here we describe some implementation details for predicting long range future with short video tracks.

\begin{table*}
	\centering
	\begin{minipage}[t]{0.48\linewidth}
		\vspace{0pt}
		\centering
		\small{
			\begin{tabular}{cccccc}
				\toprule
				& \multicolumn{5}{c}{Human3.6M \tabspace Reconst. $\downarrow$}\\
				\cmidrule(l){2-6}
				Action & 1 & 5 & 10 & 20 & 30 \\
				\midrule
				Directions &  54.5 &  57.1 &  59.2 &  60.6 &  63.3\\
				Discussion &  58.3 &  60.0 &  61.1 &  61.8 &  64.8\\
				Eating &  50.1 &  51.7 &  53.7 &  54.7 &  56.9\\
				Greeting &  60.4 &  63.9 &  68.0 &  68.1 &  71.4\\
				Phoning &  60.6 &  61.7 &  62.6 &  63.7 &  66.7\\
				Posing &  52.8 &  55.3 &  57.1 &  58.5 &  61.7\\
				Purchases &  52.5 &  54.0 &  55.8 &  56.3 &  60.0\\
				Sitting &  61.1 &  62.0 &  62.7 &  63.3 &  66.2\\
				Sitting Down &  68.3 &  69.3 &  70.0 &  71.3 &  75.0\\
				Smoking &  57.1 &  58.3 &  59.5 &  60.3 &  63.1\\
				Taking Photo &  73.3 &  74.7 &  75.9 &  77.6 &  81.9\\
				Waiting &  53.7 &  55.4 &  57.3 &  58.3 &  61.0\\
				Walk Together &  53.2 &  55.0 &  56.1 &  57.3 &  59.7\\
				Walking &  44.5 &  47.3 &  48.9 &  48.2 &  50.6\\
				Walking Dog &  62.0 &  63.8 &  65.8 &  67.7 &  70.3\\
				All &  57.7 &  59.5 &  61.1 &  62.1 &  65.1\\
				
				\bottomrule
			\end{tabular}
		}
		\caption{\small{{\bf Per-action evaluation of autoregressive future prediction in the latent space on Human3.6M with Dynamic Time Warping (DTW).} For each action, we evaluate the mean reconstruction error in \textit{mm} after applying Dynamic Time Warping for the predictions. Each column corresponds to a different number of frames into the future. We find that our model most accurately predicts periodic motion (\eg \textit{Walking}), performs reasonably on actions with small motion (\eg \textit{Eating}, \textit{Posing}, \textit{Waiting}), but less so with intricate poses (\eg \textit{Sitting Down}, \textit{Taking Photo}).}}
		\vspace{-3mm}
		\label{tab:h36m_sequences_dtw}
	\end{minipage}
	\hspace{5mm}
	\begin{minipage}[t]{0.48\linewidth}
		\vspace{0pt}
		\centering
		\small{
			\begin{tabular}{cccccc}
				\toprule
				& \multicolumn{5}{c}{Penn Action \tabspace PCK $\uparrow$}\\
				\cmidrule(l){2-6}
				Action & 1 & 5 & 10 & 20 & 30 \\
				\midrule
				Baseball Pitch &  83.9 &  81.2 &  78.2 &  75.5 &  72.1\\
				Baseball Swing &  93.6 &  92.3 &  90.2 &  92.0 &  90.8\\
				Bench Press &  63.0 &  62.9 &  62.6 &  62.6 &  62.5\\
				Bowl &  74.4 &  69.6 &  69.1 &  69.5 &  70.3\\
				Clean And Jerk &  90.8 &  89.4 &  88.9 &  88.9 &  87.9\\
				Golf Swing &  90.6 &  90.8 &  88.8 &  87.5 &  87.0\\
				Jump Rope &  93.0 &  92.8 &  92.7 &  93.0 &  92.9\\
				Pullup &  87.1 &  86.8 &  87.0 &  87.9 &  87.3\\
				Pushup &  71.4 &  71.0 &  70.6 &  71.5 &  72.2\\
				Situp &  67.4 &  66.6 &  65.8 &  66.5 &  65.4\\
				Squat &  80.8 &  80.7 &  80.4 &  78.9 &  79.1\\
				Strum Guitar &  77.8 &  78.4 &  79.2 &  78.8 &  78.7\\
				Tennis Forehand &  92.4 &  89.9 &  87.2 &  86.1 &  81.4\\
				Tennis Serve &  87.0 &  85.5 &  82.9 &  79.1 &  74.3\\
				All &  81.2 &  80.0 &  79.0 &  78.2 &  77.2\\
				
				\bottomrule
			\end{tabular}
		}
		\caption{\small{{\bf Per-action evaluation of autoregressive future prediction in the latent space on Penn Action with Dynamic Time Warping (DTW).} For each action, we evaluate the Percentage of Correct Keypoints after applying Dynamic Time Warping for the predictions. Each column corresponds to a different number of frames into the future. Note the \textit{Jumping Jacks} action category is omitted because the corresponding video sequences are too short to evaluate. On the fast sequences, our model performs more accurately on linear motion (\eg \textit{Baseball Swing}, \textit{Tennis Forehand}) than sequences that require changes in direction (\eg windups in \textit{Baseball Pitch} and \textit{Bowl}). For actions in which motion is slow, our model performance is dependent on the viewpoint quality. For instance, \textit{Jump Rope} and \textit{Clean and Jerk} tend to have frontal angles whereas \textit{Bench Press} and \textit{Situp} are often viewed from side angles that have self-occlusions. }}
		\vspace{-3mm}
		\label{tab:penn_sequences_dtw}
	\end{minipage}
	
\end{table*}

The length of consistent tracklets of human detections is limited given that people often walk out of the frame or get occluded. In Penn Action, for instance, the median video length is 56 frames. Thus, we chose to train on videos with at least 40 frames.
Recall that to avoid drifting, we train our $f_{\text{AR}}$ on its own predictions \cite{li2018auto}.
Since $f_{\text{AR}}$ has a receptive field of 13, our model must predict 14 timesteps into the future before it is fully conditioned on its own predicted movie strips.
This is further complicated by the fact that each movie strip is also causal and has its own receptive field, again pushing back when $f_\text{AR}$ can begin its first future prediction.
In principle, the maximum number of ground truth images that $f_\text{AR}$ could be conditioned on would be one less than the sum of the receptive field of $f_{\text{AR}}$ and $f_{\text{movie}}$. For a receptive field of 13, this would be $13+13-1=25$ images. However, with tracklets that have a minimum length of 40 frames, this would leave just $40 - 25 = 15$ timesteps for future prediction.
This means that just 2 predictions would be fully conditioned on previously predicted movie strips.
To support future prediction of 25 frames with a sequence length of 40, we edge pad the first image such that $\f{AR}$ is only conditioned on 15 images. This allows us to compute losses for 25 predictions into the future, leaving enough training samples in which the past input includes previous predictions. See the illustration in Figure \ref{fig:conditoning_actual}.

\begin{table*}[ht]
	\centering
	\begin{minipage}[t]{0.48\linewidth}
		\vspace{0pt}
		\centering
		\small{
			\begin{tabular}{cccccc}
				\toprule
				& \multicolumn{5}{c}{H3.6M \tabspace Reconst. $\downarrow$}\\
				\cmidrule(l){2-6}
				Action & 1 & 5 & 10 & 20 & 30 \\
				\midrule				

				Directions &  54.5 &  59.7 &  62.0 &  64.6 &  77.4\\
				Discussion &  58.3 &  61.4 &  63.0 &  66.4 &  77.5\\
				Eating &  50.1 &  53.3 &  57.5 &  57.5 &  69.5\\
				Greeting &  60.4 &  67.7 &  74.4 &  69.1 &  94.4\\
				Phoning &  60.6 &  62.7 &  64.5 &  67.8 &  80.2\\
				Posing &  52.8 &  57.2 &  60.7 &  63.2 &  80.7\\
				Purchases &  52.5 &  55.5 &  58.4 &  64.4 &  77.5\\
				Sitting &  61.1 &  62.6 &  64.6 &  66.4 &  79.0\\
				Sitting Down &  68.3 &  69.9 &  71.9 &  76.1 &  90.9\\
				Smoking &  57.1 &  59.2 &  61.7 &  64.4 &  76.8\\
				Taking Photo &  73.3 &  76.1 &  78.3 &  83.3 &  99.2\\
				Waiting &  53.7 &  57.0 &  61.0 &  61.9 &  74.8\\
				Walk Together &  53.2 &  57.0 &  60.0 &  65.7 &  78.7\\
				Walking &  44.5 &  50.5 &  55.1 &  58.5 &  75.9\\
				Walking Dog &  62.0 &  65.4 &  70.0 &  74.5 &  81.9\\
				All &  57.7 &  61.2 &  64.4 &  67.1 &  81.1\\

				\bottomrule
			\end{tabular}
		}
		\caption{\textbf{Per-action evaluation of autoregressive future prediction in the latent space on Human3.6M without Dynamic Time Warping (DTW).} Without DTW, the reconstruction errors accumulate quickly as the sequence goes on. As with DTW, sequences with less motion (\eg \textit{Eating}, \textit{Posing}, \textit{Waiting}) are easier to predict, and sequences with intricate poses (\eg \textit{Sitting Down}, \textit{Taking Photo}) are challenging.
			Note that periodic motions (\eg \textit{Walking}) are much better analyzed with DTW, which accounts for uncertainties in speed such as stride frequency. This helps account for the gap in performance without DTW.
		}
		\label{tab:h36m_sequences_nodtw}
	\end{minipage}
	\hspace{5mm}
	\begin{minipage}[t]{0.48\linewidth}
		\vspace{0pt}
		\centering
		\small{
			\begin{tabular}{cccccc}
				\toprule
				& \multicolumn{5}{c}{Penn Action \tabspace PCK $\uparrow$}\\
				\cmidrule(l){2-6}
				Action & 1 & 5 & 10 & 20 & 30 \\
				\midrule
				
				Baseball Pitch &  83.9 &  73.6 &  62.7 &  53.8 &  39.2\\
				Baseball Swing &  93.6 &  88.2 &  77.2 &  78.4 &  68.3\\
				Bench Press &  63.0 &  62.1 &  61.0 &  61.4 &  56.9\\
				Bowl &  74.4 &  63.9 &  61.3 &  59.0 &  55.5\\
				Clean And Jerk &  90.8 &  88.2 &  86.8 &  85.6 &  78.1\\
				Golf Swing &  90.6 &  86.4 &  77.3 &  61.8 &  64.6\\
				Jump Rope &  93.0 &  90.7 &  91.2 &  89.4 &  88.3\\
				Pullup &  87.1 &  85.5 &  84.4 &  85.4 &  77.5\\
				Pushup &  71.4 &  69.8 &  67.9 &  67.0 &  60.9\\
				Situp &  67.4 &  63.6 &  57.2 &  53.4 &  42.8\\
				Squat &  80.8 &  81.5 &  80.0 &  77.1 &  72.7\\
				Strum Guitar &  77.8 &  78.4 &  79.8 &  78.7 &  75.8\\
				Tennis Forehand &  92.4 &  85.7 &  76.9 &  75.1 &  50.2\\
				Tennis Serve &  87.0 &  80.9 &  71.3 &  57.7 &  40.1\\
				All &  81.2 &  77.2 &  72.4 &  67.9 &  60.1\\

				\bottomrule
			\end{tabular}
		}
		\caption{\textbf{Per-action evaluation of autoregressive future prediction in the latent space on Penn Action without Dynamic Time Warping (DTW).}
			The prediction accuracy of actions with fast motion (\eg \textit{Baseball Pitch}, \textit{Golf Swing}, \textit{Tennis Serve}, etc.) deteriorates quickly since the speed is challenging to predict. In addition, these sequences often begin with little motion as the player prepares to begin the action or is waiting for the ball to come to them. In such cases, mis-timing the start of the action results in a large quantitative penalty. As with DTW, for the slower sequences, we observe that the actions that tend to have clearer viewpoints (\eg \textit{Jump Rope}, \textit{Pullup}) outperform those that tend to be recorded from the side (\eg \textit{Bench Press}, \textit{Pushup}, \textit{Situp}).}
		\label{tab:penn_sequences_nodtw}
	\end{minipage}

\end{table*}

\begin{table*}[!b]
    
	\centering
	\small{
	\begin{tabular}{ccccccccccc}
		\toprule
		& \multicolumn{5}{c}{H3.6M \tabspace Reconst. $\downarrow$} & \multicolumn{5}{c}{Penn Action \tabspace PCK $\uparrow$}  \\
		\cmidrule(lr){2-6} \cmidrule(lr){7-11} 
		Method & 1 & 5 & 10 & 20 & 30 & 1 & 5 & 10 & 20 & 30 \\
		\midrule
		
		AR on $\Phi$ & 57.7 & \textbf{61.2} & \textbf{64.4} & 67.1 & \textbf{81.1} & \textbf{81.2} &\textbf{ 77.2} & \textbf{72.4} & \textbf{67.9} & \textbf{60.1}  \\ 
		AR on $\Phi$, no $\loss{movie strip}$ & \textbf{56.9} & \textbf{61.2} & 64.9 & \textbf{66.8} & 83.6 & 80.4 & 75.4 & 70.2 & 65.6 & 59.0\\ 
        AR on $\Theta$ & 57.8 & 65.9 & 75.9 & 91.9 & 105.2 & 79.9 & 67.8 & 56.2 & 43.4 & 35.1\\ 
        Constant & 59.7 & 71.4 & 85.9 & 101.4 & 102.8 & 78.3 & 65.5 & 54.6 & 42.3 & 32.7\\ 
        Nearest Neighbor & 90.3	 & 99.8 & 	110.3 & 	124.7 & 	133.3 & 	62.5 & 	57.6	 & 53.7	 & 44.6	 & 41.1\\
        \bottomrule
	\end{tabular}
	}
	\caption{\small{{\bf Comparison of autoregressive predictions with various baselines without Dynamic Time Warping.} We evaluate our model with autoregressive prediction in the movie strip latent space $\Phi$ (AR on $\Phi$), an ablation in the latent space without the distillation loss (AR on $\Phi$, No $\loss{movie strip}$), and predictions in the pose space $\Theta$ (AR on $\Theta$). We also show the results of the no-motion baseline (Constant) and Nearest Neighbors (NN). The performance of all methods deteriorates more quickly without Dynamic Time Warping. Our method using autoregressive predictions in the latent space still significantly outperforms the baselines.}}
	\label{tab:all_baselines_nodtw}
\end{table*}

